\documentclass[MSc]{icldt}
\usepackage{amsthm}
\usepackage{hyperref}
\hypersetup{
    colorlinks,
    citecolor=black,
    filecolor=black,
    linkcolor=black,
    urlcolor=black
}
\usepackage[british]{babel}
\usepackage{hyphenat}
\usepackage{graphicx}
\usepackage{amsmath}
\usepackage{algorithmicx}
\usepackage{algpseudocode}
\usepackage[chapter]{algorithm}
\usepackage{svg}
\usepackage{tikz}
\usetikzlibrary{positioning,shapes,arrows}
\usepackage{xcolor,colortbl}
\usepackage{longtable}
\usepackage[toc,page]{appendix}

\title{Deep Convolutional Neural Networks for Smile Recognition}
\author{Patrick Oliver GLAUNER}

\date{September 2015}
\department{Computing}
\field{Computing (Machine Learning)}
\supervisor{Professor Maja PANTIC \\ and Dr. Stavros PETRIDIS}
\declaration{I herewith certify that all material in this report which is not my own work has been properly acknowledged.}

\begin{document}

\maketitle

\begin{abstract}
This thesis describes the design and implementation of a smile detector based on deep convolutional neural networks. It starts with a summary of neural networks, the difficulties of training them and new training methods, such as Restricted Boltzmann Machines or autoencoders. It then provides a literature review of convolutional neural networks and recurrent neural networks. In order to select databases for smile recognition, comprehensive statistics of databases popular in the field of facial expression recognition were generated and are summarized in this thesis. It then proposes a model for smile detection, of which the main part is implemented. The experimental results are discussed in this thesis and justified based on a comprehensive model selection performed. All experiments were run on a Tesla K40c GPU benefiting from a speedup of up to factor 10 over the computations on a CPU.
A smile detection test accuracy of 99.45\% is achieved for the Denver Intensity of Spontaneous Facial Action (DISFA) database, significantly outperforming existing approaches with accuracies ranging from 65.55\% to 79.67\%. This experiment is re-run under various variations, such as retaining less neutral images or only the low or high intensities, of which the results are extensively compared.
\end{abstract}

\dedication{First and foremost I offer my sincerest gratitude to my supervisor Dr. Stavros PETRIDIS who has supported me throughout my thesis with his enthusiasm, patience and expertise.

I would also like to thank Professor Maja PANTIC for her passion, setting the direction of this thesis and valuable regular feedback.

Furthermore, I am eternally obliged to the feedback and advice on neural networks from Professor Sinisa TODOROVIC.}
\makededication

\tableofcontents
\listoftables
\listoffigures
\listofalgorithms

\chapter{Introduction}
\label{chapter:intro}
Neural networks have been popular in the machine learning community since the 1980s with repeating rises and falls of popularity. Their main benefit is their ability to learn complex, non-linear hypotheses from data without the need of modeling complex features. This makes them of particular interest for computer vision, in which feature description is a long-standing and largely non-understood topic. Neural networks are difficult to train and for the last ten years they have come to enormous fame under the topic "deep learning".
New advances in training methods and the movement of training from CPUs to GPUs allow to train more reliable models much faster. Deep neural networks are not a silver bullet, as training is still heavily based on model selection and experimentation.
Overall, significant progress in machine learning and pattern recognition has been made in natural language processing, computer vision and audio processing.
Leading IT companies have made significant investments into deep learning for these reasons, such as Baidu, Google, Facebook and Microsoft.

Concretely, previous work of the author on deep learning for facial expression recognition in \cite{glauner_ISO} resulted in a deep neural network model that significantly outperformed the best contribution to the 2013 Kaggle facial expression competition \cite{kaggle_competition}.
Therefore, a further investigation on the recognition of action units and in particular smile using deep neural networks and convolutional neural networks seems desirable. Only very few works on this topic have been reported so far, such as in \cite{gudi}.
It would also be interesting to compare the input of the entire face versus the mouth to study differences in the performance of deep convolutional models.

\chapter{Background report: neural networks}
\label{chapter:nn}
This chapter provides an overview of different types of neural networks, their capabilities and training challenges, based on \cite{glauner_ISO}. This chapter does not provide an introduction to neural networks, the reader is therefore referred to \cite{bishop} and \cite{mitchell} for a comprehensive introduction to neural neural networks.

Neural networks are inspired by the brain and composed of multiple layers of logistic regression units, called neurons.
They experienced different periods of hypes in the 1960s and 1980s/90s.
Neural networks are known to be able to learn complex hypotheses for regression and classification. Conversely, training neural networks is difficult, as their cost functions have many local minima.
Hence, training tends to converge to a local minimum, resulting in poor generalization of the network.
For the last ten years, neural networks have been celebrating a comeback under the term deep learning, taking advantage of many hidden layers in order to build more powerful machine learning algorithms.

\section{Feed-forward neural networks}
\begin{figure}[h!]
    \centering
    \includegraphics[width=0.6\textwidth]{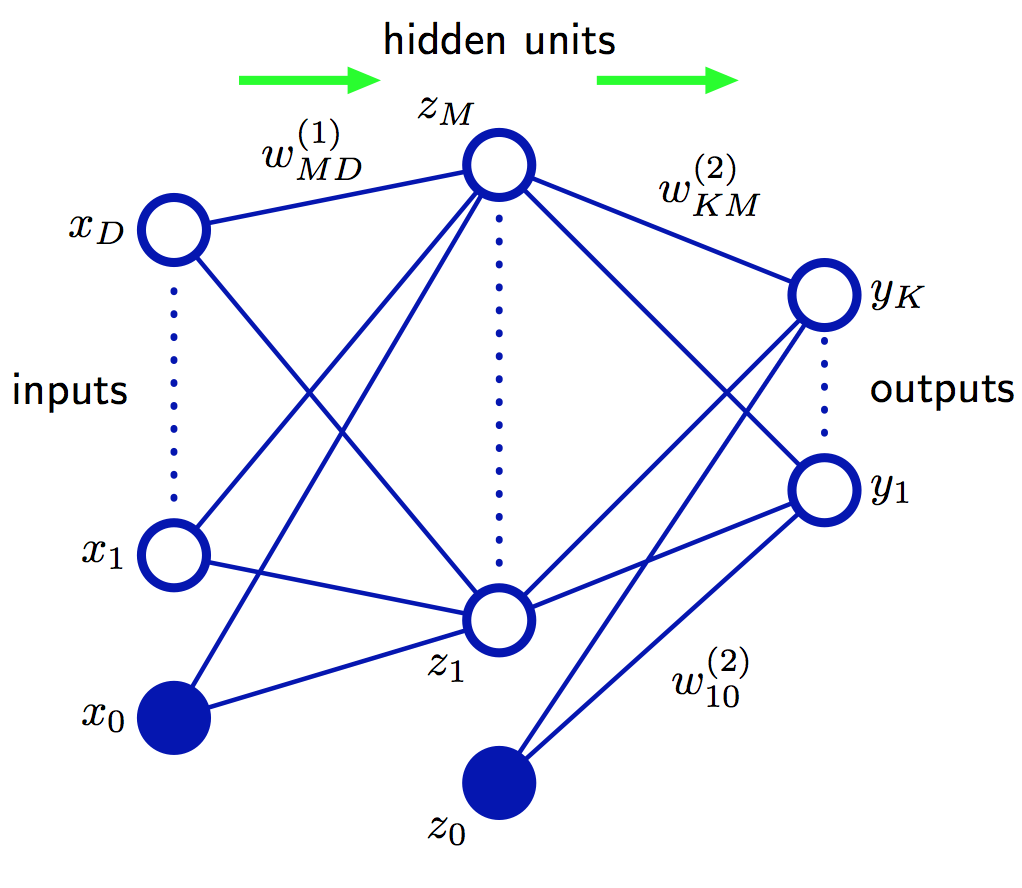}
    \caption{Neural network with two input and output units and one hidden layer with two units and bias units $x_0$ and $z_0$ \cite{bishop}.}
    \label{fig:NN}
\end{figure}

Feed-forward neural networks are the simplest type of neural networks. They are composed of an input layer, one or more hidden layers and an output layer, as visualized in Figure~\ref{fig:NN}.

Using learned weights $\Theta$ or $W$, they propagate an input through the network to the output to make predictions.
The activation of unit $i$ of layer $j+1$ can be calculated as follows:
\begin{align}
\label{eq:activation}
z_i^{(j+1)} = \sum_{k=0}^{s_{j}} \Theta_{ik}^{(j)}x_k \\
a_i^{(j+1)} = g\left(z_i^{(j+1)}\right)
\end{align}

$g$ is an activation function, for which often the Sigmoid activation function $\frac{1}{1 + e^{-x}}$ is used in the hidden layers. The Sigmoid function or its generalization, the softmax function, are used for classification problems in the output layer units.
For regression problems, the sum of Equation~\ref{eq:activation} is used directly in the output layer without the use of any activation functions.

In order to learn the weights, a cost function is minimized. There are different cost functions, such as the least squares or cross-entropy cost function, described in \cite{mitchell}.
The latter one has been reported to generalize better and speed up learning as discussed in \cite{ng_mlc}.

\subsection{Difficulty of training}
In order to learn the weights, Algorithm~\ref{alg:backpropagation} named backpropagation is used to efficiently compute the partial derivatives, which are then fed into an optimization algorithm, such as gradient descent (Algorithm~\ref{alg:gradientdescent}) or stochastic gradient descent (Algorithm~\ref{alg:stochasticgradientdescent}), as described in \cite{lecun_research}. Those three algorithms are based on \cite{ng_mlc}.

\begin{algorithm}[H]
\caption{Backpropagation: training size $m$.}
\label{alg:backpropagation}
\begin{algorithmic}
\State $\Theta_{ij}^{(l)} \gets rand(-\epsilon, \epsilon)$ (for all $l,i,j$)
\State $\Delta_{ij}^{(l)} \gets 0$ (for all $l,i,j$)
\For{$i=1$ to $m$}
\State $a^{(1)} \gets x^{(i)}$
\State Perform forward propagation to compute $a^{(l)}$ for $l=2,3,...,L$
\State Using $y^{(i)}$, compute $\delta^{(L)}=a^{(L)}-y^{(i)}$ \Comment {"error"}
\State Compute $\delta^{(L-1)},\delta^{(L-2)},...,\delta^{(2)}$: $\delta^{(l)}=(\Theta^{(l)})^T\delta^{(l+1)}\circ g'(z^{(l)})$
\State $\Delta^{(l)} \gets \Delta^{(l)}+\delta^{(l+1)}(a^{(l)})^T$  \Comment {Matrix of errors for units of a layer}
\EndFor
\State $\frac{\partial}{\partial \Theta_{ij}^{(l)}}J(\Theta) \gets \frac{1}{m}\Delta_{ij}^{(l)}$
\end{algorithmic}
\end{algorithm}

\begin{algorithm}[H]
\caption{Batch gradient descent: training size $m$, learning rate $\alpha$.}
\label{alg:gradientdescent}
\begin{algorithmic}
\Repeat
\State $\theta_j  \gets \theta_j - \alpha \frac{\partial}{\partial \theta_j}J(\theta)$ (simultaneously for all $j$)
\Until{convergence}
\end{algorithmic}
\end{algorithm}

\begin{algorithm}
\caption{Stochastic gradient descent: training size $m$, learning rate $\alpha$.}
\label{alg:stochasticgradientdescent}
\begin{algorithmic}
\State Randomly shuffle data set
\Repeat
\For{$i=1$ to $m$}
\State $\theta_j  \gets \theta_j - \alpha \frac{\partial}{\partial \theta_j}J(\theta, (x^{(i)}, y^{(i)}))$ (simultaneously for all $j$)
\EndFor
\Until{convergence}
\end{algorithmic}
\end{algorithm}

Generally, the more units in a neural network, the higher its expressional complexity. In contrast, the more units, the more it tends to overfit.
To prevent overfitting, various approaches have been described in the literature, including $L_1$/$L_2$ regularization \cite{ng_l1_l2}, early stopping, tangent propagation \cite{bishop} and dropout \cite{dropout_simple}.

\section{Deep neural networks}
\begin{figure}[h!]
    \centering
    \includegraphics[width=0.9\textwidth]{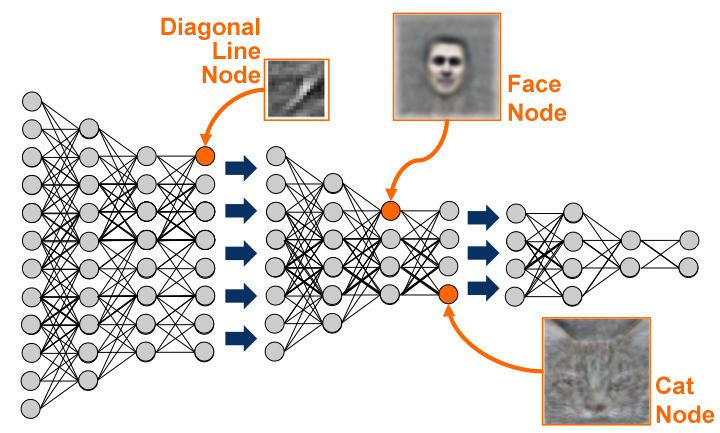}
    \caption{Deep neural network layers learning complex feature hierarchies \cite{cat_visualization}.}
    \label{fig:NN_example}
\end{figure}

Deep neural networks use many hidden layers. This allows to learn increasingly more complex features hierarchies, as visualized in Figure~\ref{fig:NN_example} for the Google Brain \cite{google_brain}. Such architectures are of enormous benefit, as the long-standing problem of feature description in signal processing disappears to a large extend.

Conversely, training of deep neural networks gets more difficult because of the increased number of parameters. As described in \cite{difficulty_training} and \cite{pretraining_help}, backpropagation does not scale to deep neural networks: starting with small random initial weights, the backpropagated partial derivatives go towards zero. As a result, training becomes infeasible and is called the vanishing gradient problem.

\subsection{Training methods}
For deep neural networks, training has therefore been split in two parts: pre-training and fine-tuning.
Pre-training allows to initialize the weights to a location in the cost function which can be optimized quickly using regular backpropagation.

Various pre-training methods have been described in the literature. Most prominently, unsupervised methods, such as Restricted Boltzmann Machines (RBM) in \cite{hinton_guide} and \cite{hinton_science} or autoencoders in \cite{ng_tutorial} and \cite{deng} are used.
Both methods learn exactly one hidden layer. This hidden layer is then used as input to the next RBM or autoencoder to learn the next hidden layer. This process can be repeated for many times in order to pre-train a so-called Deep Belief Network (DBN) or Stacked Autoencoder, composed of RBMs or autoencoders respectively. In addition, there are denoising autoencoders defined in \cite{stacked_denoising}, which are autoencoders that are trained to denoise corrupted inputs.
Furthermore, other methods such as discriminative pre-training \cite{hinton_speech} or reduction of internal covariance shift \cite{covariance_shift} have been reported as effective training methods for deep neural networks.

\subsection{Activation functions}
\label{chapter:activation}

\begin{figure}[h!]
    \centering
    \includegraphics[width=0.7\textwidth]{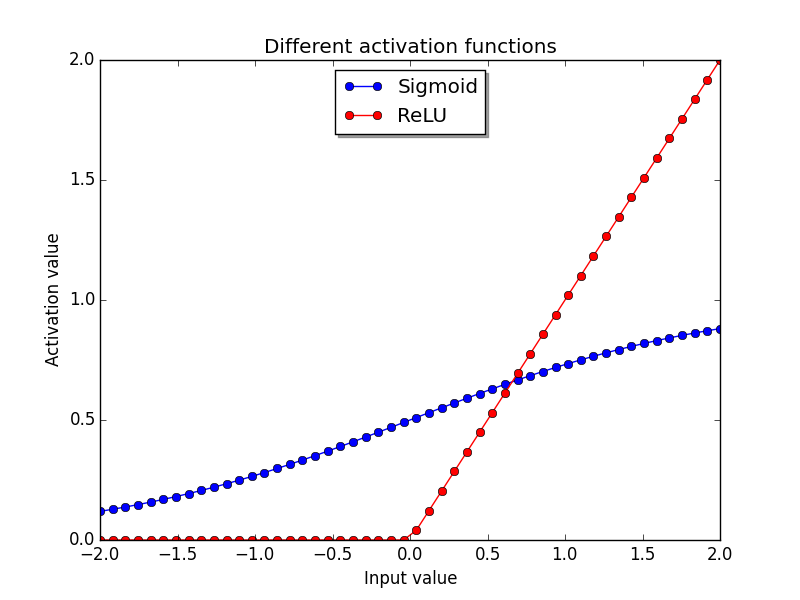}
    \caption{Sigmoid and ReLU activation functions.}
    \label{fig:activation_functions}
\end{figure}

In the past, mostly Sigmoid units have been used in the hidden layers, with Sigmoid or linear units in the output layer for classification or regression, respectively. For classification, the softmax activation is preferred in the output layer. As described by Norvig in \cite{norvig}, the output of a set unit is much stronger than the others. Another benefit of softmax is that it is always differentiable for a weight.
Recently, the so-called rectified linear unit (ReLU) has been proposed in \cite{rectified}, which has been used successfully in many deep learning applications.
Figure~\ref{fig:activation_functions} visualizes the Sigmoid and ReLU functions.

ReLU has a number of advantages over Sigmoid, reported in \cite{rectified} and \cite{glorot}.
First, it is much easier to compute as it is either $0$ or the input value.
Also, Sigmoid has for non-activated input values less than or equal to $0$ an activation value of greater than $0$. In contrast, ReLU models biological behavior of neurons more accurately, as it is $0$ for those cases.
With many units set to $0$, a sparse activation of the networks follows, which is another form of regularization.
Furthermore, the vanishing gradient problem becomes less of an issue as ReLU units result in a simpler cost function.
Last, for some experiments, ReLU reduces the importance of pre-training or may not be necessary at all.

\subsection{Application to facial expression data}
In the context of this project, deep neural networks have been successfully applied to facial expression recognition in \cite{glauner_ISO}. In that study, RBMs, autoencoders and denoising autoencoders were compared on a noisy dataset from a 2013 Kaggle  challenge named "Emotion and identity detection from face images" \cite{kaggle_competition}. This challenge was won by a neural network presented in \cite{tang}, which achieved an error rate of 52.977\%.
In \cite{glauner_ISO}, a stacked autoencoder was trained with an error of 39.75\%. In a subsequent project, this error could be reduced further to 28\% with a stacked denoising autoencoder \cite{glauner_ocado}.
This study also showed that deep neural networks are a promising machine learning method for this context, but not a silver bullet as data pre-processing and intensive model selection are still required.

\section{Recurrent neural networks}
\label{chapter:RNN}
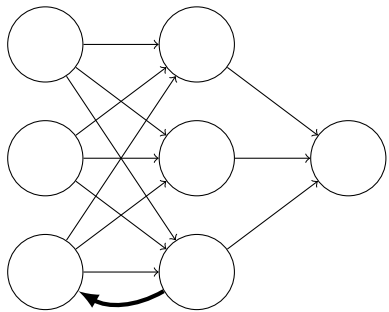
\begin{figure}[h!]
\centering
\begin{tikzpicture}[
  node distance=1cm and 1cm,
  minimum size=1cm,
  mynode/.style={draw,circle,align=center}
]
\node[mynode] (A12) {};
\node[mynode,below=0.5cm of A12] (A22) {};
\node[mynode,below=0.5cm of A22] (A32) {};
\node[mynode,left=0cm and 1cm of A12] (X1) {};
\node[mynode,left=0cm and 1cm of A22] (X2) {};
\node[mynode,left=0cm and 1cm of A32] (X3) {};
\node[mynode,right=0cm and 1cm of A22] (L3) {};
\path (X1) edge[->] (A12)
(X1) edge[->] (A22)
(X1) edge[->] (A32)
(X2) edge[->] (A12)
(X2) edge[->] (A22)
(X2) edge[->] (A32)
(X3) edge[->] (A12)
(X3) edge[->] (A22)
(X3) edge[->] (A32)
(A12) edge[->] (L3)
(A22) edge[->] (L3)
(A32) edge[->] (L3)
(A32) edge [-latex,bend left,ultra thick] node {} (X3);
\end{tikzpicture}
\caption{Simple recurrent neural network with one recurrent connection from the hidden layer to the input layer in \textbf{bold}.}
\label{figure:recurrentneuralnetwork}
\end{figure}

Recurrent neural networks (RNNs) are cyclic graphs of neurons as displayed in Figure~\ref{figure:recurrentneuralnetwork}.

They have increased representational power as they create an internal state of the network which allows them to exhibit dynamic temporal behavior.
Training RNNs is more complex as this depends on their structure. The RNN in Figure~\ref{figure:recurrentneuralnetwork} can be trained using a simple variant of backpropagation.
In practice, recurrent networks are more difficult to train than feedforward networks and do not generalize as reliably.

\subsection{Long short-term memory}
\begin{figure}[h!]
    \centering
    \includegraphics[width=0.7\textwidth]{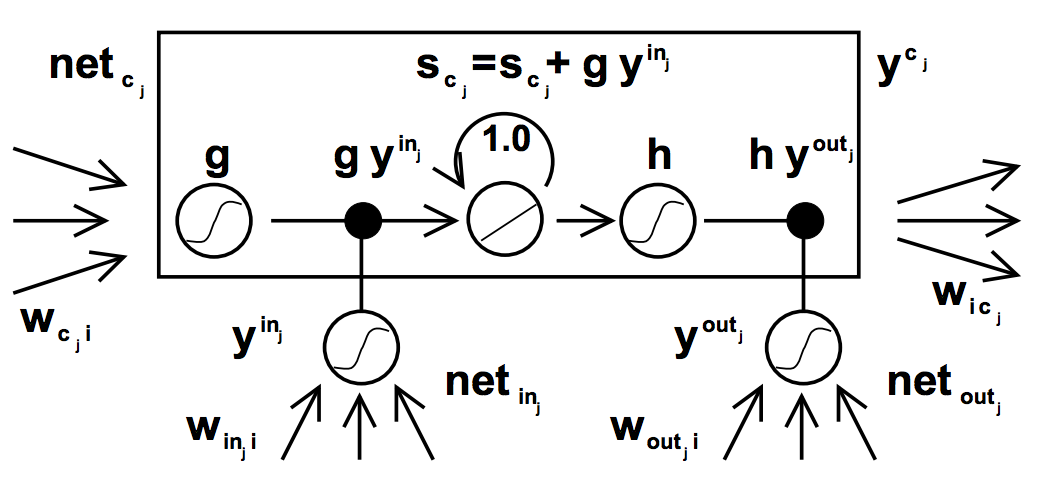}
    \caption{LSTM cell: the integral sign stands for the Sigmoid function, the large filled dot for a multiplication \cite{LSTM}.}
    \label{fig:LSTM_cell}
\end{figure}

A long short-term memory (LSTM) defined in \cite{LSTM} is a modular recurrent neural network composed of LSTM cells. A LSTM cell is visualized in Figure~\ref{fig:LSTM_cell}.

\begin{figure}[h!]
    \centering
    \includegraphics[width=0.9\textwidth]{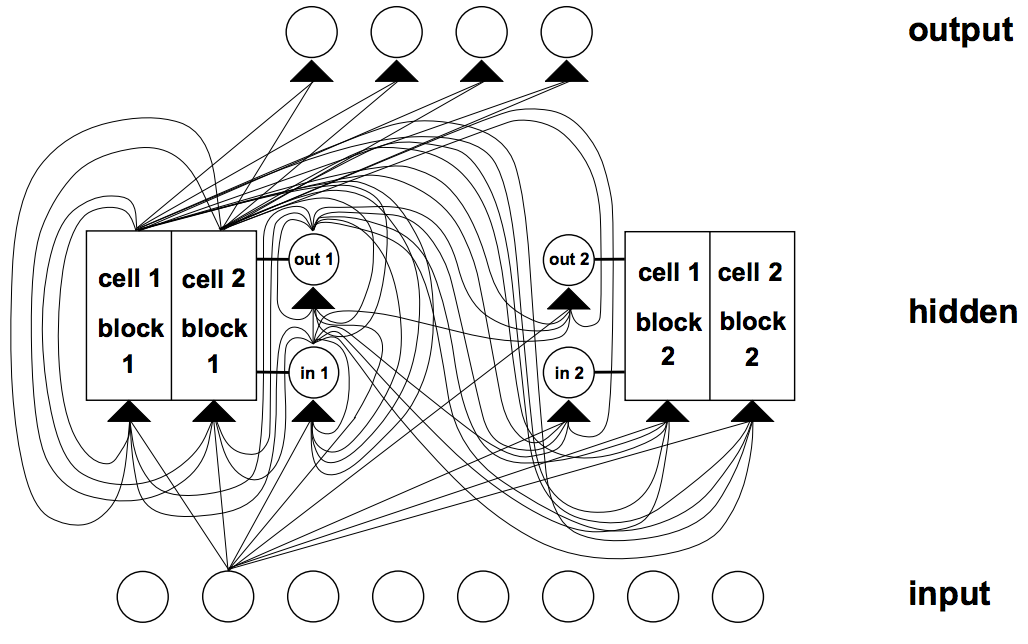}
    \caption{Example LSTM network: eight input units, four output units, and two memory cell blocks of size two \cite{LSTM}.}
    \label{fig:LSTM_example}
\end{figure}

Inputs $w_{c{_j}i}$ are fed in, for which a value $g$ is computed using the Sigmoid function of the dot product of the input and weights.
The second Sigmoid unit $y^{in_j}$ is the input gate.
If its output value is near to zero, the product $g\cdot y^{in_j}$ is near to zero, too, thus zeroing out the input value. As a consequence, this blocks the input value, preventing it from going further into the cell.
The third Sigmoid unit $y^{out_j}$ is the output gate.
Its function is to determine when to output the internal state of the cell. This is the case when the output of this Sigmoid unit is close to one.
LSTM cells can be put together in a modular structure, as visualized in Figure~\ref{fig:LSTM_example} to build complex recurrent neural networks.

Training LSTMs takes advantage of backpropagation through time, a variant of backpropagation. Its goal is to minimize the LSTM's total cost on a training set. LSTMs have been reported to outperform regular RNNs and Hidden Markov Models in classification and time series prediction tasks.
LSTMs have also been reported in \cite {LSTM_video} to perform well on prediction of image sequences.

\section{Convolutional neural networks}
Invariance to transformations is a desired property of learning algorithms. Typical variances of images and videos include translation, rotation and scaling. Tangent propagation \cite{bishop} is one method in neural networks to handle transformations by penalizing the amount of distortion in the cost function.
Convolutional neural networks (CNNs) are a different approach to implementing invariance in neural networks, which are inspired by biological processes. CNNs were initially proposed by LeCun in \cite{MNIST_LeNet}. They have been successfully applied to computer vision problems, such as hand-written digit recognition.

In images, nearby pixels are strongly correlated, a property of which local features take advantage of. In a hierarchical approach, local features are used in the first stage of pattern recognition, allowing recognition of more complex features.

\begin{figure}[h!]
    \centering
    \includegraphics[width=0.7\textwidth]{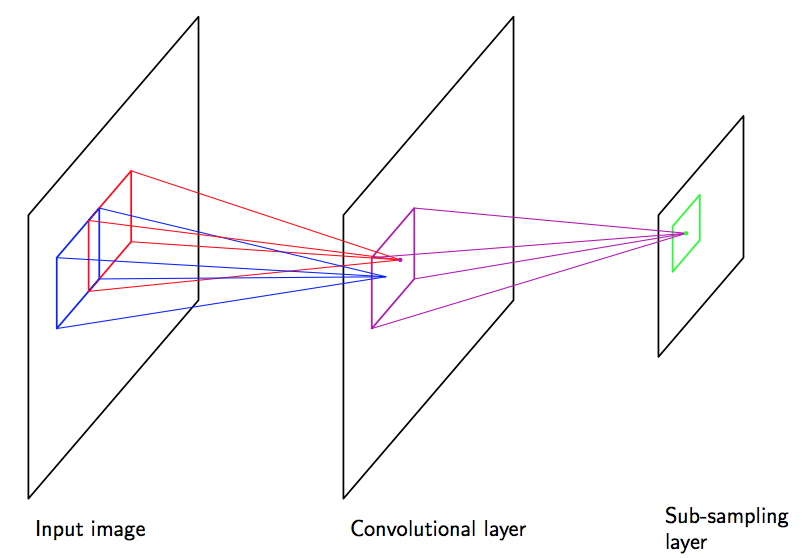}
    \caption{Illustration of a convolutional neural network \cite{bishop}.}
    \label{fig:CNN_example}
\end{figure}

The concept of CNNs is illustrated in Figure~\ref{fig:CNN_example} for a layer of convolutional units, followed by a sub-sampling layer, as described in \cite{bishop}.

The convolutional layer is composed of so-called feature maps. Units in a feature map take inputs from a small subregion of the input.
All units in a feature map share the same weights, which is called weight sharing. Replicating units in this way allows for features to be detected independently of their position in the visual field.

The subsampling layer takes small regions of convolutional layer as input and computes the average (or maximum or other functions) of those inputs, multiplied by a weight and finally applies the Sigmoid function to the value.
The result of a unit in the subsampling layer is relatively insensitive to small shifts or rotations of the image in the corresponding regions of the input space.
This concept can be repeated for more times to subsequently be more invariant and to detect more complex features.

Because of the constraints of weights, the number of independent parameters in the network is smaller than in a fully-connected network. This allows to train the network faster and to be less prone to overfitting.
Training of CNNs requires minimization of a cost function. The idea of backpropagation can be applied to CNN with a small modification taking into account the weight sharing.

\section{Processing of image sequences}
\label{chapter:nn:sequences}
\begin{figure}[h!]
    \centering
    \includegraphics[width=0.9\textwidth]{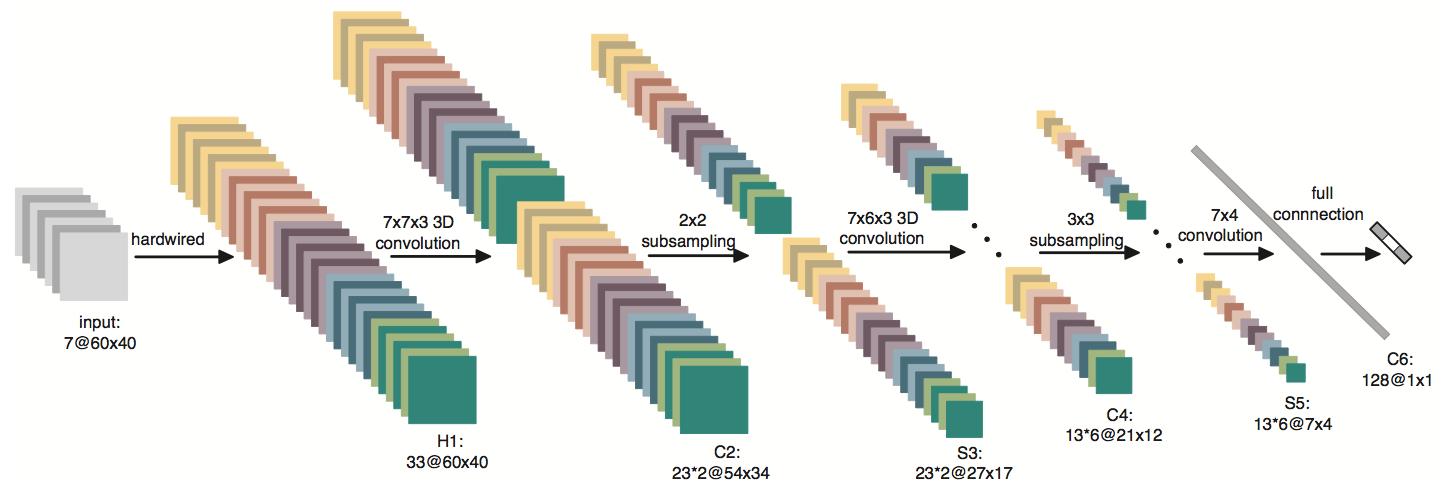}
    \caption{Multiple convolutions to process video input \cite{karpathy}.}
    \label{fig:multiple_convolutions}
\end{figure}

Recently, CNNs have been reported to work well on processing of image sequences, for example in \cite{karpathy} for multiple convolutions, as visualized in Figure~\ref{fig:multiple_convolutions}. 

A related approach is reported in \cite{ji}.
CNNs are expanded to work on image sequences instead of single images. The extra weights need to be initialized in a way so that training can easily optimize them. An extensive study and comparison of different initialization methods is provided in \cite{initialization}.

\begin{figure}[h!]
    \centering
    \includegraphics[width=0.4\textwidth]{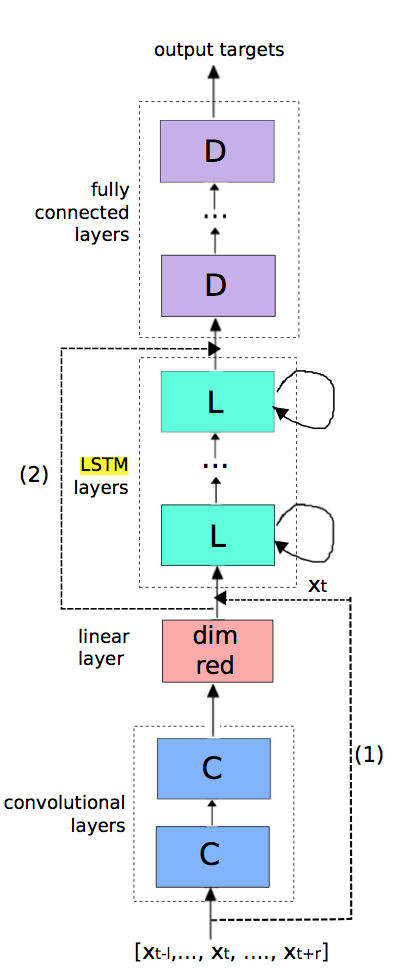}
    \caption{Deep neural network composed of convolutions, LSTMs, dimensionality reduction and regular layers \cite{sainath}.}
    \label{fig:full_architecture}
\end{figure}

\cite{sainath} describes a deep architecture composed of convolutions, LSTMs and regular layers for a NLP problem. It begins with multiple convolutional layers. Next, a linear layers follows with fewer units in order to reduce the dimensionality of the features recognized by the convolutional layers. Next, the reduced features are fed into a LSTM. The output of the LSTM  is then used in regular layers for classification.
The entire architecture is visualized in Figure~\ref{fig:full_architecture}.

\begin{figure}[h!]
    \centering
    \includegraphics[width=0.9\textwidth]{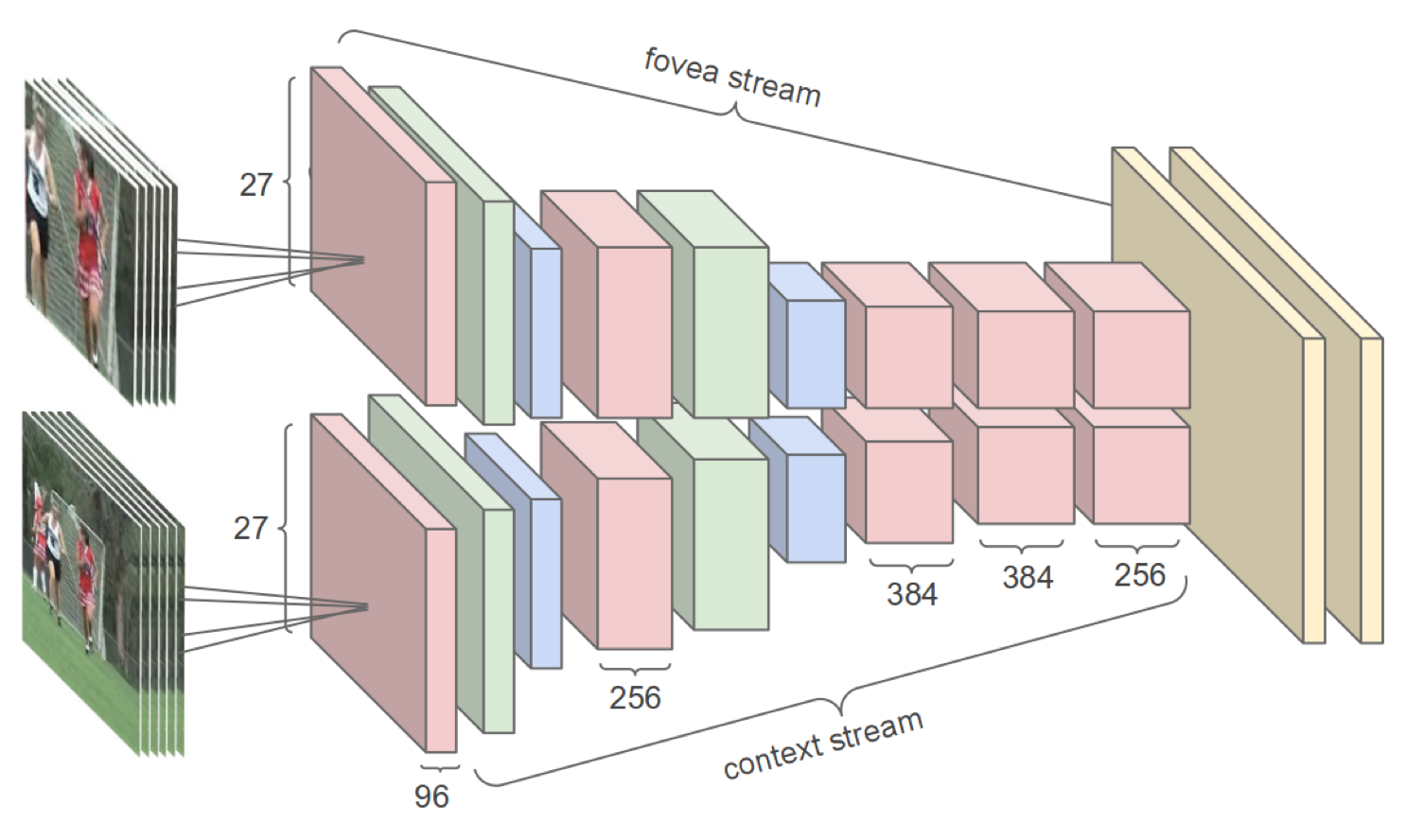}
    \caption{Fusion of low-resolution with higher-resolution of the center of the video \cite{karpathy}.}
    \label{fig:fusion_center}
\end{figure}

Similar architectures exist for processing of image sequences and are elaborated further.
Very successful results using fusion of different video inputs have been reported, too. For example, a reported architecture in \cite{karpathy} fuses a low-resolution version of the input with a higher-resolution input of the center of the video. This is visualized in Figure~\ref{fig:fusion_center}.

\begin{figure}[h!]
    \centering
    \includegraphics[width=0.9\textwidth]{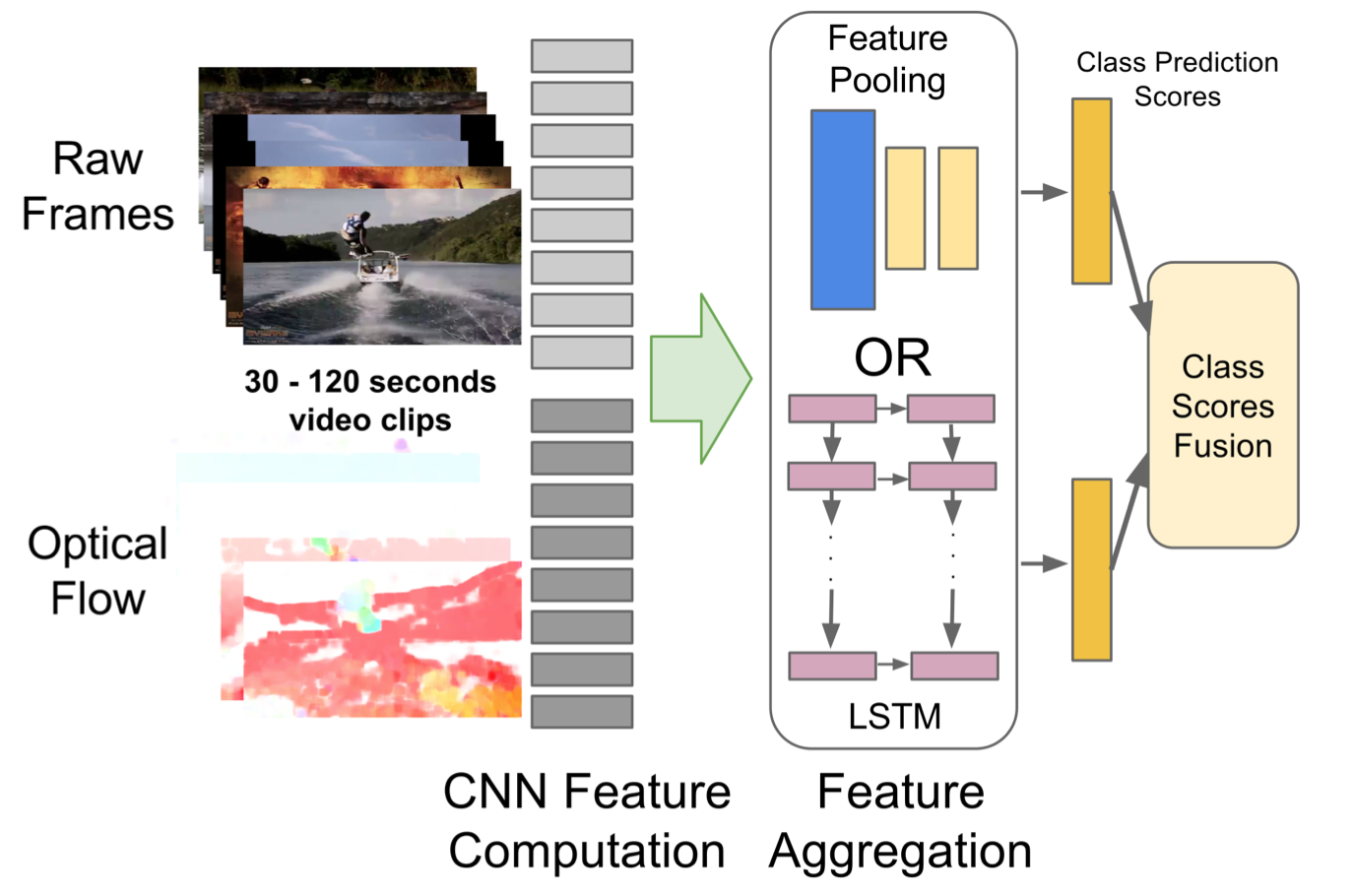}
    \caption{Fusion of low-resolution with optical flow \cite{beyond}.}
    \label{fig:fusion_flow}
\end{figure}

Conversely, \cite{beyond} fuses a low-resolution version of the input with the optical flow, as visualized in Figure~\ref{fig:fusion_flow}.

\begin{figure}[h!]
    \centering
    \includegraphics[width=0.9\textwidth]{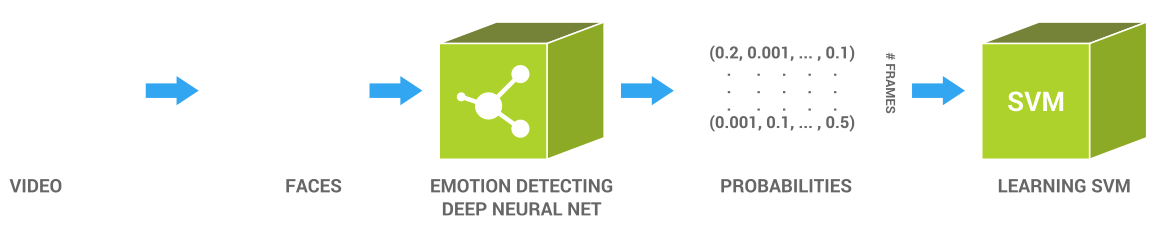}
    \caption{Final stage done by SVM instead of neural network \cite{kahou}.}
    \label{fig:final_svm}
\end{figure}

The final stage of video classification can alternatively be done by a different classification, such as a Support Vector Machine (SVM). This is described in \cite{kahou} and visualized in Figure~\ref{fig:final_svm}.

Furthermore, a spatio-temporal convolutional sparse autoencoder for sequence classification is described in \cite{baccouche}.

\chapter{Selection of databases}
\label{chapter:databases}
In this chapter, various popular databases relevant to action unit recognition are presented. Each database includes annotations per frame of the respective action units, among other features. Furthermore, statistics of the distribution of action units were generated for each database in order to select databases rich of smiles.

\section{FACS coding}
The Facial Action Coding System (FACS) is a system to taxonomize any facial expression of a human being by their appearance on the face. It was published by Paul Ekman and Wallace V. Friesen in 1978 \cite{FACS}. Relevant to this thesis are so-called Action Units (AUs), which are the basic actions of individual facial muscles or groups of muscles. Action units are either set or unset. If set, different levels of intensity are possible.

\section{Available databases}
Popular databases in the field of action unit recognition and studies of facial expressions include the following, which are presented briefly in this section. The reader is referred to the relevant literature for details.

The Affectiva-MIT Facial Expression Dataset (AMFED) \cite{AMFED} contains 242 facial videos (168,359 frames), which were recorded in the wild (real world conditions).
The Chinese Academy of Sciences Micro-expression (CASME) \cite{CASME} database was filmed at 60fps and contains 195 micro-expressions of 22 male and 13 female participants.
The Denver Intensity of Spontaneous Facial Action (DISFA) \cite{DISFA} database contains videos of 15 male and 12 female subjects of different ethnicities. Action unit annotations are on different levels of intensity.
The Geneva Multimodal Emotion Portrayals (GEMEP) \cite{GEMEP} contains audio and video recordings of 10 actors which portray 18 affective states.
The MAHNOB Laughter \cite{MAHNOB} database contains 22 subjects recorded using a video camera, a thermal camera and two microphones. Recorded were laughter, posed smiles, posed laughter and speech. It includes 180 sessions with a total duration of 3h and 49min.

\begin{figure}[h!]
    \centering
    \includegraphics[width=0.4\textwidth]{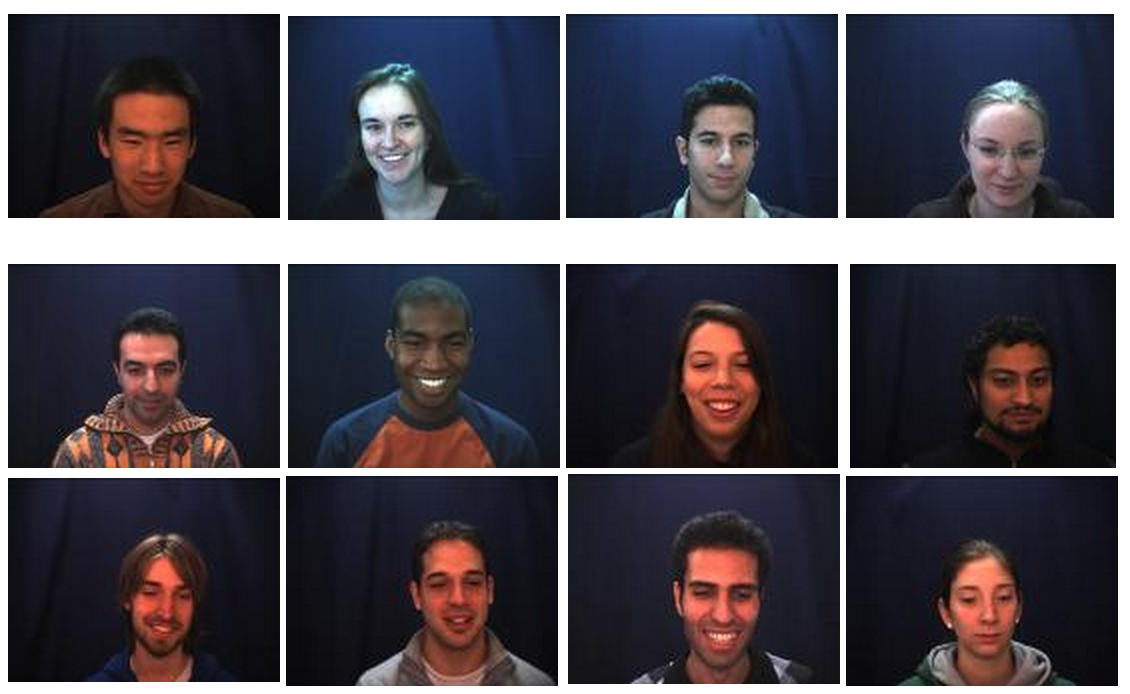}
    \caption{Sample images of the DISFA database \cite{DISFA}.}
    \label{fig:DISFA}
\end{figure}

The UNBC-McMaster Shoulder Pain Expression Archive Database \cite{PAIN} contains 200 video sequences of participants that were suffering from shoulder pain and their corresponding spontaneous facial expressions. In total, it includes 48,398 FACS coded frames.

\section{Distribution of action unit intensities}
\begin{figure}[h!]
    \centering
    \includegraphics[width=0.3\textwidth]{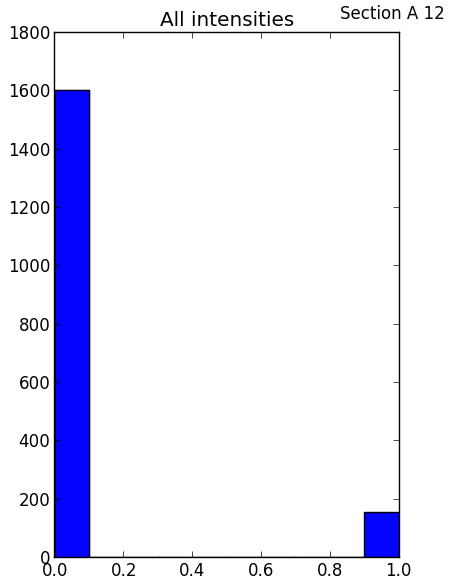}
    \caption{Binary statistics of CASME database.}
    \label{fig:CASME_STAT}
\end{figure}

For the databases presented in the previous section, statistics of the annotations of action units were generated. This task has proven to be complex, as the structure of each database is different and need to be parsed accordingly\footnote{Without the use of an abstract programming language like Python, this task alone would have been easily an entire thesis project on its own.}. Comprehensive plots and statistics of the individual action units were generated. For example, Figure~\ref{fig:CASME_STAT} represents the binary distribution of AU12, which represents smile in FACS coding, of the CASME database.

\begin{figure}[h!]
    \centering
    \includegraphics[width=0.6\textwidth]{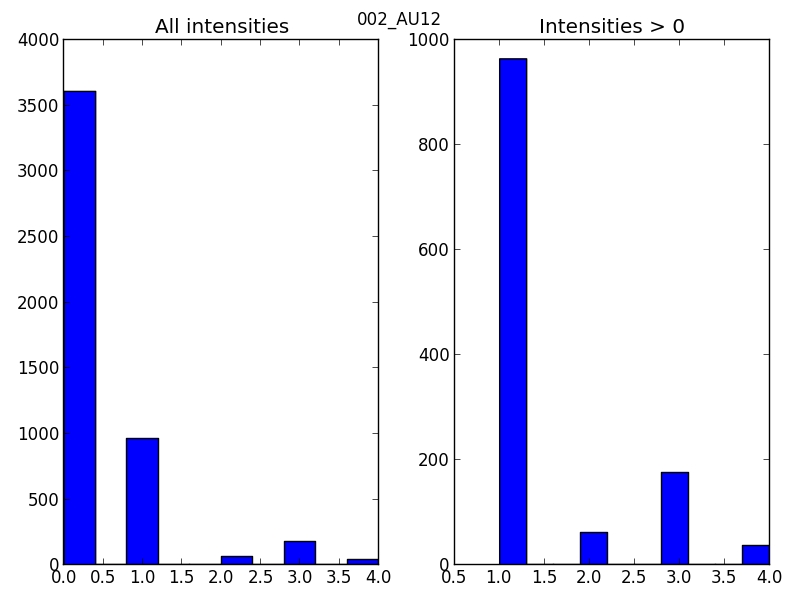}
    \caption{Intensity statistics for video 002 of DISFA database. Left subplot: all intensities, right subplot: all positive intensities.}
    \label{fig:DISFA_STAT1}
\end{figure}

Statistics were generated at different levels of granularity. For example, Figure~\ref{fig:DISFA_STAT1} contains the multi-valued intensity distribution of AU12 of video 002 of the DISFA database.

\begin{figure}[h!]
    \centering
    \includegraphics[width=0.6\textwidth]{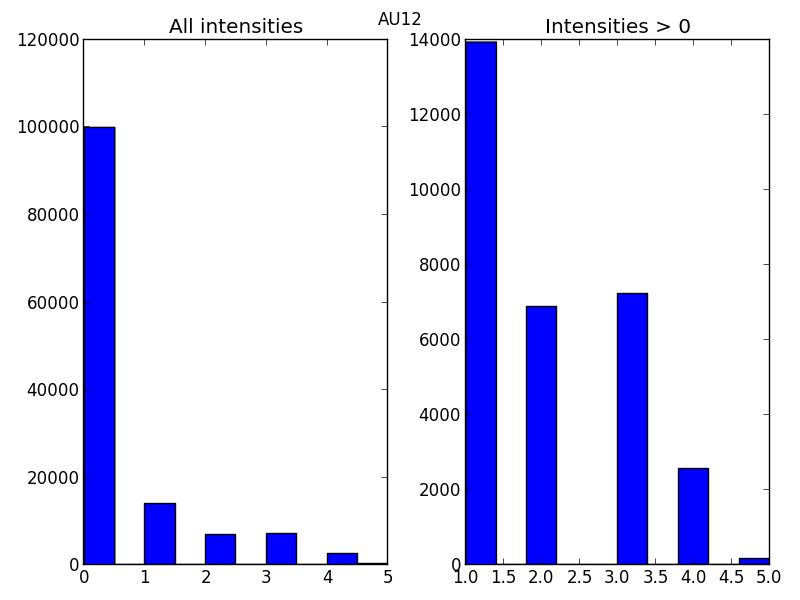}
    \caption{Intensity statistics for all videos of DISFA database. Left subplot: all intensities, right subplot: all positive intensities.}
    \label{fig:DISFA_STAT2}
\end{figure}

Conversely, Figure~\ref{fig:DISFA_STAT2} contains the multi-valued intensity distribution of AU12 of the entire DISFA database.

Table~\ref{table:some_stats} contains a selection of action units of the different databases. Due to different terminology, the AMFED database does not use AU12, but a feature called "smile" as explained in \cite{AMFED}.

\begin{table}[h!]
\centering
\begin{tabular}{ c || c | c | c | c | c | c }
  & AMFED & CASME & DISFA & GEMEP & MAHNOB Laughter & Shoulder Pain \\
\hline
\hline
AU1 & - & 1976 & 8778 & 1584 & - & - \\
\hline
AU12 & - & 264 & 30794 & 2692 & - & 6887 \\
\hline
AU16 & - & 126 & - & 310 & - & - \\
\hline
AU21 & - & - & - & 95 & - & - \\
\hline
Laughter & - & - & - & - & 6404 & - \\
\hline
Smile & 77062 & - & - & - & - & - \\
\hline
negAU12 & 350 & - & - & - & - & - \\
\end{tabular}
\caption{Selected statistics of action units in databases: an integer denotes the number of frames in which an action unit is set (intensity $> 0$). A hyphen indicates that an action unit is not available in a database.}
\label{table:some_stats}
\end{table}

The full statistics of all action units are available in Appendix~\ref{chapter:fullstats}.

\section{Selected databases}
In order to be selected for the following experiments, a database-AU pair must satisfy two conditions: First, the action unit should be sufficiently often set in the annotations of a database in order to be better learnable. Second, the database images should be available in an aligned format. Aligned images are cropped, retaining the actual face in its center, plus the availability of facial landmark point annotations.

\subsection{DISFA}
\label{chapter:DISFA}
\begin{figure}[h!]
    \centering
    \includegraphics[width=0.3\textwidth]{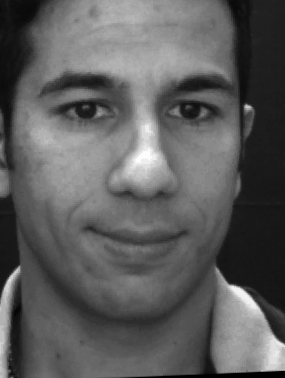}
    \caption{Sample image of aligned DISFA database of size $285\times 378$ pixels \cite{DISFA}.}
    \label{fig:DISFA_ALIGNED}
\end{figure}

For these reasons, the DISFA database was selected to be used for smile recognition.
The aligned version consists of 27 videos of 4845 frames each, with 130,815 images in total. Each image is $285\times 378$ pixels on a grey-value scale. Figure~\ref{fig:DISFA_ALIGNED} is a sample image of the aligned version of DISFA.

As one video in DISFA lacks the 4845th image and in order to avoid handling this edge case, only the first 4844 images of every video have been used. In total, 130,788 images are used. For those images, further statistics have been generated. In particular, 30,792 have AU12 set. Table~\ref{table:disfa_intensities} contains the distribution of AU12. In total, 82,176 images have some action unit(s) set and 48,612 images have no action units set at all.

\begin{table}[h!]
\centering
\begin{tabular}{c || c }
Intensity & Count \\
\hline
\hline
0 & 99996\\
1 & 13942\\
2 & 6868\\
3 & 7233\\
4 & 2577\\
5 & 172
\end{tabular}
\caption{Distribution of AU12 in DISFA.}
\label{table:disfa_intensities}
\end{table}

In the original paper on DISFA \cite{DISFA}, multi-class SVMs were trained for the different levels 0-5 of action unit intensity. Test accuracies for the individual levels and for the binary action unit recognition problem are reported for three different feature description techniques.
In those three cases, binary accuracies of 65.55\%, 72.94\% and 79.67\% are reported.

\subsection{Others}
For the same reasons, the shoulder pain database is of further interest of smile detection for further experiments, such as a multi-database smile detector.
Furthermore, the laughter in the MAHNOB Laughter database may be of interest in future experiments, as laughter includes smile.
AMFED was not considered further, as "smile" is not AU12, but something slightly different, but may be of interest in further experiments, too.

\chapter{Model}
\label{chapter:model}
The goal of this project is to recognize and predict action units from videos, in particular smiles. A regular deep neural network would not suit this task for two main reasons: First, deep neural networks do not support handling translation or other distortions of the input, which happen frequently in facial videos. Second, deep feed-forward neural networks do not have a state, therefore making processing of videos difficult as they require handling of states in order to recognize or predict action units.
In this chapter, the proposed model for smile detection is explained in detail, of which the first part is implemented.
In order to train it in a reasonable amount of time, a powerful underlying computing infrastructure has been used.

\section{Proposed model}
Based on findings described in Chapter~\ref{chapter:nn:sequences}, an initial model has been defined and refined after discussions with other experts, including Sinisa Todorovic \cite{sinisa}. The model can be summarized as follows: Feature extraction in the first stage, followed by the temporal part.

For feature extraction, a CNN is trained on images of the entire face or an area suitable for smile detection, such as the mouth. This CNN is followed by one or multiple layers of a regular (dense) neural network for discrimination of the features. The exact architecture of the network, such as the number of convolutions, number of hidden layers, etc. is subject to model selection, which was extensively performed in Chapter~\ref{chapter:resevaluation}. The size of the input is also subject to model selection as one input unit is needed per input pixel. The larger the input image, the better, as more data and details are available. Conversely, the model becomes more complex and more difficult to train, with overfitting or long training time as possible consequences.

The output of this network is fed into the second part, which handles temporal relationships. There are different possibilities how to model it.
On the one hand, state-of-the-art methods, such as Hidden Markov Models (HMMs), could be used. On the other hand, recurrent neural networks are of particular interest for this project. As described in Chapter~\ref{chapter:RNN}, LSTMs are reported to perform well on temporal data and are known to be able to outperform HMMs. Therefore, LSTMs are chosen for this part, followed by one or multiple layers of a regular neural network for discrimination of the features.

\begin{figure}[h!]
    \centering
    \includegraphics[width=\textwidth]{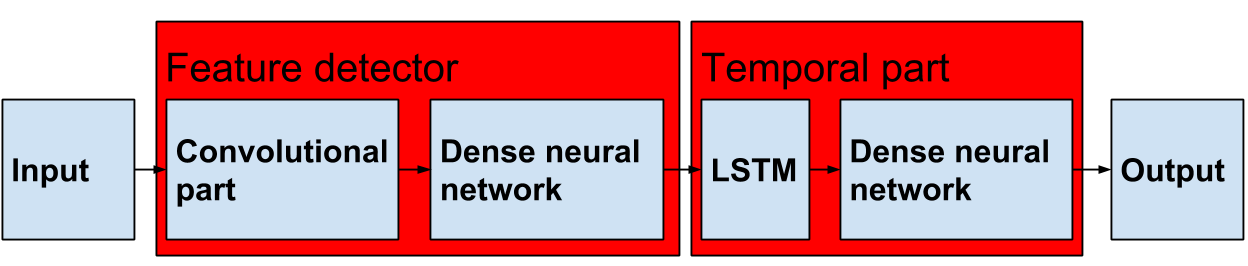}
    \caption{Proposed model.}
    \label{fig:model}
\end{figure}

The proposed model is visualized in Figure~\ref{fig:model}.

In the literature, related experiments on other databases have been performed. Results were reported, in which the two parts were subsequently trained, i.e. the feature extraction was trained first and used to train the temporal part \cite{beyond} \cite{beyond_temporal}.
In contrast, other models that were trained end-to-end are described in the literature, too \cite{beyond_temporal} \cite{graves} \cite{hannun}. An end-to-end trained model seems preferable for those experiments and would therefore also be interesting to investigate.

\section{Implementation}
\label{chapter:implementation}
In this section, the key implementation decisions made are described and reasoned. In the course of this thesis, the underlying algorithms of deep learning were not be implemented due to time constraints. Therefore, suitable libraries were selected and the results of this evaluation are explained in this section.

\subsection{Selection of deep learning library}
In \cite{glauner_ISO}, the MATLAB Deep Learning Toolbox \cite{deep_toolbox} has been proven to be easy and quick to use for deep learning experiments. It also supports the training of CNNs, but tends to be slow for many parameters and large datasets.

To speed up training, the use of GPUs is highly preferred. The two main libraries in this domain are Theano \cite{theano} and Caffe \cite{caffe}. Both take advantage of GPUs for computations and have been extensively compared for this project and the results are summarized in this section.
Theano is a general purpose numerical computation library for Python. Its instructions are run either on the CPU or can be compiled to GPU code without any necessary knowledge about GPUs. It does not offer deep learning functionality natively, but allows to write efficient deep learning methods.
Caffe is a deep learning framework implemented in C++ with integrations for Python and MATLAB. It requires a developer to simply specify a deep learning architecture. Working on deep learning is possible on an abstract level. Nonetheless, when modifications are necessary or new models are to be built, they have to be implemented in C++, which requires a deep understanding of the Caffe architecture.

Keeping this future flexibility in mind is important as proposed in the outreach in Chapter~\ref{chapter:conclusions}. Therefore, Theano appears to be beneficial to be used, as Python is more abstract and the implementation of the model will not require an understanding of the underlying library architecture.

There are multiple deep learning libraries that build on top of Theano, such as Blocks and Fuel or Lasagne. Both are still under heavy development, which make a qualitative comparison difficult and the results may be subject to change in the near future for newer versions.
Blocks and Fuel are two different libraries. Blocks is a deep learning library that supports CNNs, RNNs and LSTMs. Its input comes from a Fuel source, which is a data stream framework primarily built to support Blocks. During the evaluation, it appears to be powerful and abstract, but over-engineered and difficult to use.
Lasagne is a simpler library for deep neural networks and CNNs, which is easier to use. In contrast, it lacks support of LSTMs.

Considering the benefits and drawbacks of the respective libraries, Lasagne was chosen for the implementation of the model. As Lasagne lacks support of LSTMs, a separate LSTM library was chosen, as described in the following section.

\subsection{Selection of LSTM library}
There is an extension of Lasagne for LSTMs \cite{github_LASAGNE_LSTM} which prove to be effective in the evaluation. It is most straightforward to use together with the feature detector of the first stage. Also, an end-to-end training of the entire model is possible using this library. Nonetheless, the project has only one main committer coming with uncertainty if it will be kept in sync with Lasagne in the future.

Support for use of GPUs for training is also offered by CURRENNT \cite{sourceforge_CURRENNT}, a C++ library for recurrent neural networks. No support for Python is offered by this library, making integration into existing code of the feature detector more difficult.

Furthermore, RNNLIB \cite{sourceforge_RNNLIB} is a popular library for recurrent neural networks, including LSTMs. Its Python wrapper allows easy integration in existing code of the feature detector. It lacks support of GPUs, which may come with long training time for the large database of this project.

Based on these considerations, the Lasagne LSTM extension seems most preferable because of the same data format, functions and easy integration into existing code.

\subsection{Progress of implementation}
As mentioned previously, Lasagne is still under development, which proved to make the implementation of the model more time consuming than initially expected due to changes in the API. In particular, a lot of demo code did not work correctly, leaving the author of this thesis with unexpected behavior and no useful error messages.

Once these issues were sorted out, the implementation of the training and model selection of the feature detector in Chapter~\ref{chapter:resevaluation} was straightforward due to the abstraction provided by Lasagne. 

In the course of this project, only the first stage of the model, the feature detector, was implemented. Due to time constraints, the second part could not be implemented.
Because of the overall high test accuracies of the feature detector in Chapter~\ref{chapter:resevaluation}, there is also a lesser need of adding temporal capabilities to this model at this point.

\section{Computing infrastructure}
\begin{figure}[h!]
    \centering
    \includegraphics[width=\textwidth]{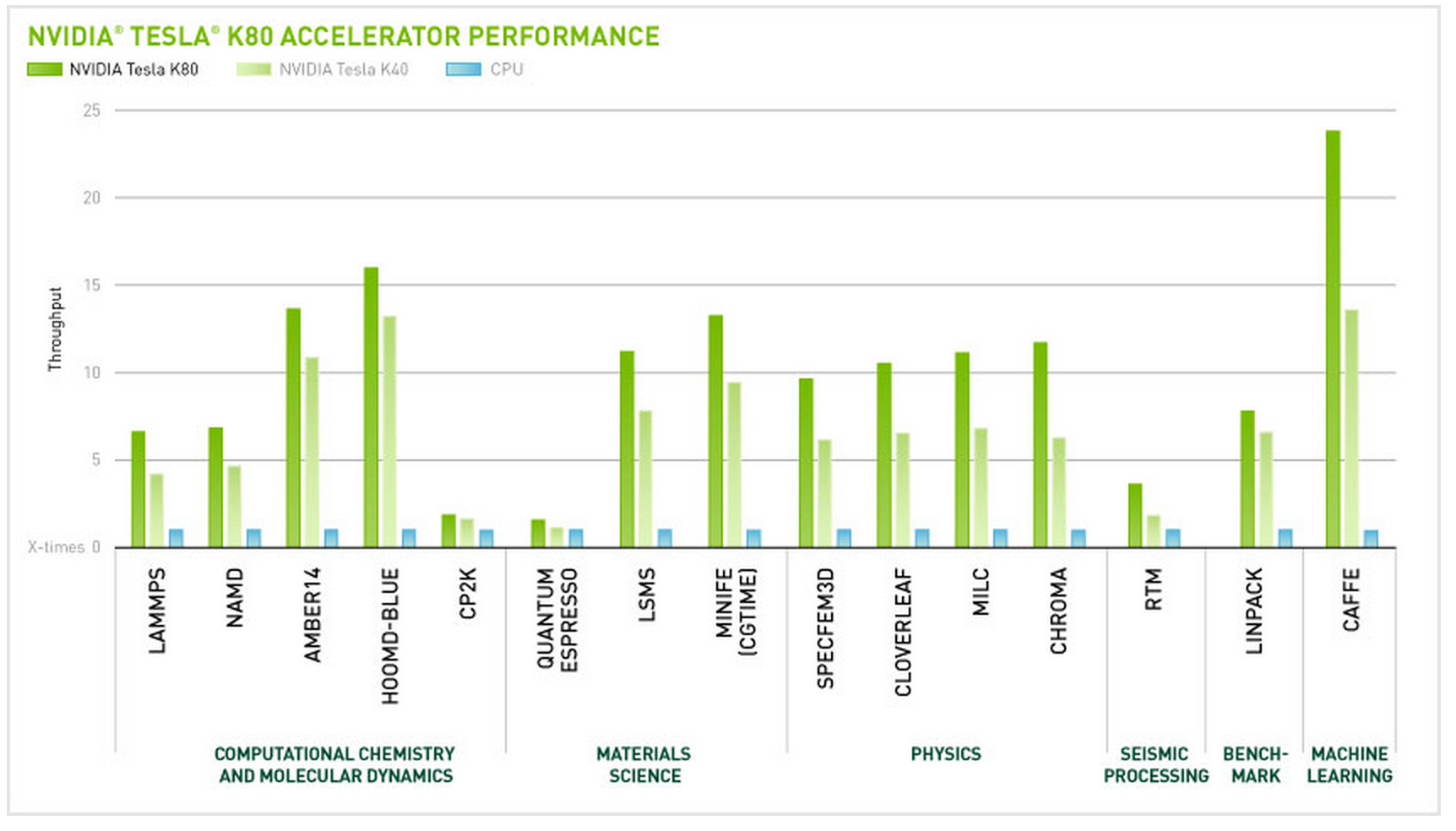}
    \caption{Examples of speedups using Tesla graphic card series \cite{tesla}.}
    \label{fig:TESLA_speed}
\end{figure}

In initial experiments, GPU acceleration provided by Theano has proven to speed up the training by factor 3-10 in comparison to a CPU. The experiments of this project cannot be run on the GPU of a modern notebook, such as a latest MacBook Pro, because the provided RAM of the GPU is too small to fit some of the models.
In these experiments, various GPUs were used including a GeForce GTX TITAN Black \cite{geforce_titanx} or a even more powerful Tesla K40c \cite{tesla}. For the Tesla series, significant speedups have been measured for different applications as collected in Figure~\ref{fig:TESLA_speed}. 

\begin{figure}[h!]
    \centering
    \includegraphics[width=0.8\textwidth]{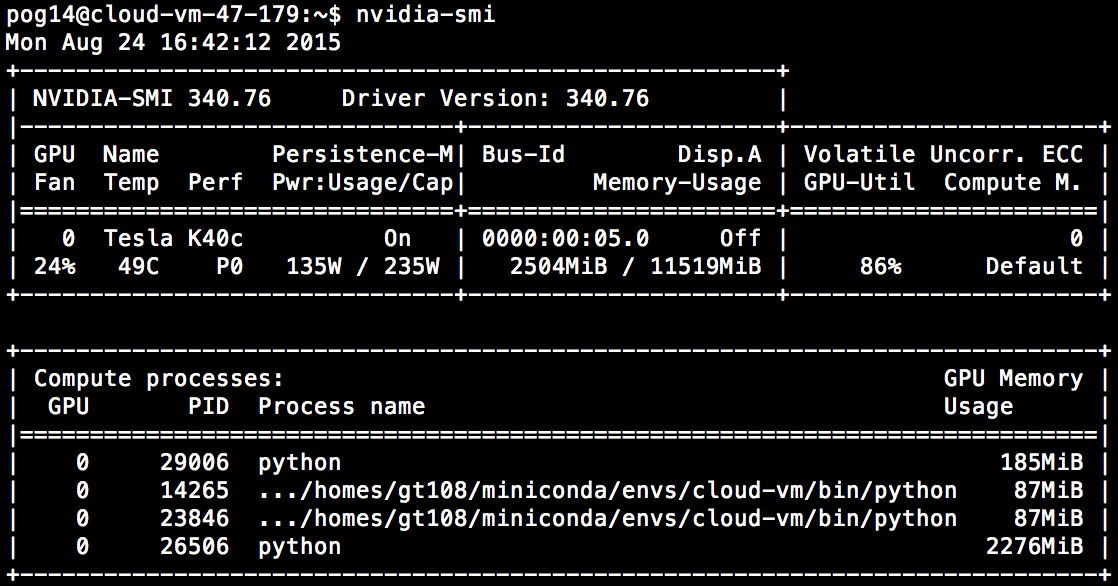}
    \caption{Output of \texttt{nvidia-smi} for sample experiment.}
    \label{fig:TESLA_screenshot}
\end{figure}

For the experiments in Chapter~\ref{chapter:resevaluation}, a server containing a Tesla K40c with 12 GB of GPU RAM and 64 GB of regular RAM was chosen. Both memories are sufficiently large to store the model and training data. The Tesla would allow to run multiple experiments at the same time, as a single experiment only uses a fraction of the GPU RAM as visualized in Figure~\ref{fig:TESLA_screenshot}.

\chapter{Towards a static convolutional smile detector}
\label{chapter:resevaluation}
In this chapter, experiments for smile detection using the convolutional feature detector are performed on the DISFA database. An essential task is model selection to pick the best architecture from a large permutation of many possible parameters. Starting with regular smile detection, only low or high intensity smiles are retained for smile recognition. Finally, low intensity smiles are discriminated against high intensity smiles.
In order to perform the experiments in time, preliminary assumptions made are reasoned.

\section{Selected parameters and assumptions}
Today, there is a lack of literature or research on neural networks for sample complexity or general rules to choose an architecture.
Therefore, in order to find good parameter values for the feature detector, model selection needs to be performed.

\subsection{Candidate parameters to be optimized}
There are many possible parameters to be optimized and reported in the literature, including:

\begin{enumerate}
\item Number of convolution-pooling pairs
\item Architecture of convolutions, such as the number of feature maps and their size
\item Architecture of poolings, such as the type of pooling, pooling size or whether to pool at all
\item Type of activation function, such as rectified linear units (ReLU), softmax or Sigmoid
\item Type of regularization, such as dropout or $L_2$
\item Number of hidden layers
\item Number of units in the hidden layers
\item Learning rate
\item Momemtum
\end{enumerate}

Parameters 1 to 3 concern the convolutional part of the network. A number of optimizations are possible, such as the number of convolution-pooling pairs and how to build the individual convolutions and poolings. The parameters to be optimized include the size and number of feature maps, the type of pooling and the pooling size. Another question is whether to use pooling at all, as good results without pooling were reported in \cite{springenberg}.

Activation functions are described in Chapter~\ref{chapter:activation} and the remaining parameters are described in \cite{glauner_ISO}. A further discussion is omitted in this part of this thesis.

\subsection{Selected parameters and values}
In order to reduce the duration of the model selection to a realistic scale, various assumptions were made. 
For convolutions and subsequent poolings, many parameters could be optimized in model selection, exploding the possible search space. Therefore, a number of parameters are fixed, based on experiments with the same library on MNIST: convolutions are for areas of $5\times 5$ pixels and in each convolutional layer, 32 feature maps are used.
Subsequent pooling is for areas of $2\times 2$ pixels and only max pooling is used, as the concrete type of pooling is reported to be less relevant in the literature \cite{beyond}.
Convolution-pooling pairs are used throughout the experiments, no single convolutions not followed by pooling \cite{springenberg} \cite{sainath}.
For reasons of simplicity, a convolution-pooling pair is simply named convolution in the remainder of this thesis.

The benefits of rectified linear (ReLU) units are discussed in Chapter~\ref{chapter:activation}. As they are reported to outperform Sigmoid units, ReLU units are used throughout all experiments. As the only exception, softmax is used in the output layer.

For regularization, dropout is the only explicit regularization method used in the model selection. $L_2$ regularization is not used at all, as a wide spectrum of possible values would have to be tested. As a consequence, model selection would take significantly more time. Furthermore, ReLU units serve as an implicit regularization method because they lead to sparse activations in the network.

The learning rate is fixed to $\alpha=0.01$ and not subject to model selection as it would also significantly prolong the model selection. The same considerations apply to the momentum, which is fixed to $\mu=0.9$. Overall, the momentum is expected to have less impact due to the use of ReLU units, as reasoned in Chapter~\ref{chapter:activation}. Both values are taken from the Lasagne MNIST showcase, for which they worked effectively.

Based on these considerations, the following parameters are subject to model selection: number of convolution-pooling pairs, number of hidden layers, number of units in hidden layers and and dropout.
Table~\ref{table:cv_values} contains the values chosen for model selection of the respective parameters and default values. The values and default values were picked, based on prior experience and initial assumptions. For the default values, the simplest values were picked, except for dropout. For dropout, $p=0.5$ is chosen in the Lasagne MNIST showcase and proved to be effective in initial bottom line experiments in Chapter~\ref{chapter:bottom}.
The table also contains in parentheses the short name chosen for parameters, which are used in subsequent tables.

\begin{table}[h!]
\centering
\begin{tabular}{ c || c | c }
Parameter & Values & Default value \\
\hline
\hline
Number of convolution-pooling pairs (\#Convs)& 1, 2, 3 & 1 \\
\hline
Number of hidden layers (\#Hidden layers) & 1, 2, 3 & 1 \\
\hline
Number of units in hidden layers (\#Units hidden layers) & 100, 200, 300, 400 & 100 \\
\hline
Dropout & 0, 0.1, 0.5, 0.7 & 0.5
\end{tabular}
\caption{Parameters and possible values used in model selection.}
\label{table:cv_values}
\end{table}

\subsection{Cost function and performance metrics}
For the following model selection, the cross-entropy loss/cost function is used for $m$ examples, hypothesis $h_{\theta}$ and target values $y^{(i)}$:
\begin{align}
J(\theta) = \frac{1}{m}\sum_{i=1}^m\left(-y^{(i)}\log(h_{\theta}(x^{(i)}))-(1-y^{(i)})\log(1-h_{\theta}(x^{(i)}))\right)
\end{align}
In contrast to other possible cost functions, such as least squares, it is known to generalize better and that training has been reported to converge faster \cite{ng_mlc}.

In the following model selection, both the cross-entropy loss and the test accuracy (classification rate for this binary problem) are output. This decision has been made because of the following reasons: the cross-entropy loss is mathematically more accurate, whereas the test accuracy is more intuitive for humans. Nonetheless, it must be noted that both metrics are different and not fully comparable.

\subsection{Input size}
\begin{figure}[h!]
    \centering
    \hfill
    \subfloat[Mouth input\label{subfig-1:dummy}]{%
        \includegraphics[width=0.3\textwidth]{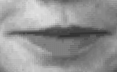}
    }%
    \hfill
    \subfloat[Face input\label{subfig-2:dummy}]{%
      \includegraphics[width=0.3\textwidth]{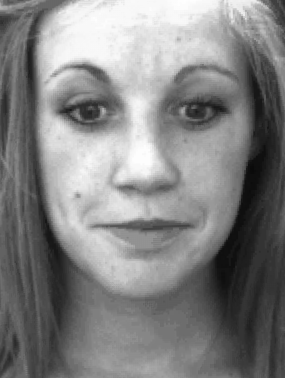}
    }%
    \hfill
    ~
    \caption{Different input parts: a) mouth, b) face \cite{DISFA}. (Not at actual input size/proportions.)}
    \label{fig:mouth_face}
\end{figure}

All experiments are run for two different sources of data: mouth or entire face in order to find out if the mouth alone is as meaningful as the face for smile detection, see Figure~\ref{fig:mouth_face}.

The aligned images are $285\times 378$ pixels, as covered in Chapter~\ref{chapter:DISFA}.

Because of facial landmarks contained in the aligned images, the location of the mouth can be computed. A bounding box to include the mouth of every image was computed.
This bounding box is of size $128\times 104$ pixels, which would be 13,312 input units. In order to reduce overfitting and to speed up training, both dimensions have been reduced to $2/3$ of their original size using a bilinear interpolation. Therefore, the mouth input is $85\times 69$ pixels, requiring 5,865 input units in total.

The same considerations have been applied to the input of the entire face. The size reduction factor is stronger than for the mouth in order to do the training in a realistic time, at the expense of possibly meaningful features. As a result, faces are fed in as $95\times 121$ pixels, requiring 11,495 input units in total.

MNIST \cite{MNIST_LeNet} is a commonly used toy problem in many deep learning publications. MNIST consists of 60,000 test examples of size $28\times 28$ pixels. Therefore, this training task is not only more complex in terms of what to detect, but also in terms of the amount of data points and number of input pixels.

\section{Bottom lines}
\label{chapter:bottom}
In order to evaluate the underlying software and hardware infrastructure, very initial experiments on much smaller inputs have been performed on fixed architectures.
Mouth images are of size $37\times 28$, whereas face images are of size $40\times 48$.
The architectures used are a regular neural network of 2 hidden layers of 800 units each, followed by a softmax output layer of 2 units. The other architecture is a convolutional neural network of two convolution and pooling stages and a fully connected hidden layer in front of the softmax output layer. Details are omitted for these initial experiments.
The test losses and accuracies are summarized in Table~\ref{table:bottom_line_before_cv}.

\begin{table}[h!]
\centering
\begin{tabular}{ c | c || c | c }
Input & Network & Test loss & Test accuracy \\
\hline
\hline
 Mouth & NN & 0.258068 & 90.16\% \\
 \hline
 Mouth & CNN & \textbf{0.167116} & \textbf{93.34\%} \\
 \hline
 \hline
 Face & NN & 0.331730 & 86.92\% \\
 \hline
 Face & CNN & \textbf{0.188780} & \textbf{92.36\%} \\
\end{tabular}
\caption{Bottom line experiments for both inputs for NN and CNN. Optimal values per input in \textbf{bold}.}
\label{table:bottom_line_before_cv}
\end{table}

For both inputs, the CNNs outperform the NNs by a margin of 3.18\% and 5.44\% for mouth and face, respectively. Both CNNs achieve an accuracy of over 90\%. Nonetheless, the NNs perform well given the noise in the the data.
The CNN for the input of the mouth outperforms the CNN for the input of the face. This cannot be generalized as the face images are too small for practical purposes and because no model selection was performed.
Nonetheless, these initial results serve as a bottom line for future experiments.

\section{Model selection for full dataset}
\label{chapter:full}
\begin{figure}[h!]
    \centering
    \includegraphics[width=0.7\textwidth]{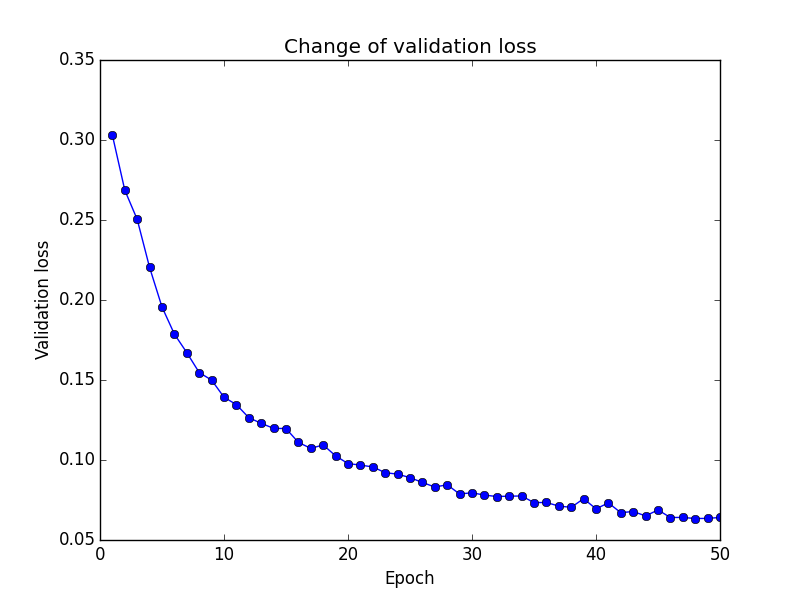}
    \caption{Change of validation loss for mouth data for 2 convolutional layers, other parameters default. For full dataset.}
    \label{fig:validation_loss_full}
\end{figure}

For the following model selection, the entire DISFA database was split in a training/validation/test ratio of 60\%/20\%/20\%. The validation set was used exclusively at each epoch for loss validation. After the respective number of epochs, the test set was used to compute test loss and test accuracy.
Each of the parameters was optimized independently using default values of the other parameters at the same time. Model selection was performed for two different types of input: mouth or the entire face and for two different number of epochs: 10 and 50 epochs.
An exhaustive search or more epochs were not possible due to the enormous training times, for which statistics of each trained permutation are available in Appendix~\ref{chapter:time}.
In most examples, the validation loss drops off quickly before converging slowly, as visualized in Figure~\ref{fig:validation_loss_full}. Therefore, at this stage of the experiment, the maximum number of epochs is set to 50.
Each experiment is performed exactly once, which comes with a certain bias, but reduces training time significantly. Chapter~\ref{chapter:repeat} provides an experimental justification that this bias is low.

\subsection{Mouth}

For the mouth input, Tables~\ref{table:cv_mouth_complete_10_full} and \ref{table:cv_mouth_complete_50_full} contain the result of the model selection for 10 and 50 epochs, respectively. The first line of each table is the combination of all default values and therefore serve as a bottom line in each experiment.
For 10 epochs, for none of the four parameters the default one was chosen to be better than the ones available in the selection. In particular, lowest losses are returned for 2 convolutions, 3 hidden layers, 400 hidden units per layer and a dropout value of 0.1.
Overall, test accuracies are on a high level and individually tested values for the parameters have minor effect as test accuracies range from $90.99\%$ to $94.54\%$. A dropout value of $0.1$ results in the highest accuracy, whereas a dropout value of $0.7$ reduces the test accuracy the most.

For 50 epochs, these selected values remain the same, except that in this case 2 hidden layers have the lowest test loss.
Overall, test accuracies are on a even higher level and individually tested values for the parameters have minor effect as test accuracies range from $94.54\%$ to $97.70\%$. A dropout value of $0.1$ results in the highest accuracy, whereas a dropout value of $0$ reduces the test accuracy the most.

\subsection{Face}

Conversely, for the entire face input, Tables~\ref{table:cv_face_complete_10_full} and \ref{table:cv_face_complete_50_full} contain the result of the model selection for 10 and 50 epochs, respectively.
For 10 epochs, the default one was chosen to be better than the ones available in the selection for the number of convolutions and the number of hidden units. In particular, lowest losses are returned for 1 convolution, 2 hidden layers, 100 hidden units per layer and a dropout value of 0.
Overall, test accuracies are on a high level and individually tested values for the parameters have minor effect as test accuracies range from $94.69\%$ to $96.36\%$. A dropout value of $0$ results in the highest accuracy, whereas 3 convolutions reduce the test accuracy the most.

For 50 epochs, these selected values remain the same for the number of convolutions and dropout. The optimal number of hidden layers and hidden units change to 1 and 400, respectively.
Overall, test accuracies are on a even higher level and individually tested values for the parameters have minor effect as test accuracies range from $97.37\%$ to $98.57\%$. A dropout value of $0$ results in the highest accuracy, whereas a dropout value of $0.7$ reduces the test accuracy the most.

\subsection{Comparison of mouth vs. face}
For both input parts, mouth and face, the model selection for 50 epochs returned different optimal parameters, which are collected in Table~\ref{table:cv_both_final_models_full}.

\begin{table}[h!]
\centering
\begin{tabular}{ c || c | c | c | c }
Input & \#Convs & \#Hidden layers & \#Units hidden layers & Dropout \\
\hline
\hline
Mouth & 2 & 2 & 400 & 0.1 \\
 \hline
Face & 1 & 1 & 400 & 0 \\
\end{tabular}
\caption{Selected parameter values for mouth and face input. For full dataset.}
\label{table:cv_both_final_models_full}
\end{table}

Overall, the test accuracies are slightly higher for the input of the entire face than just the mouth for 10 and 50 epochs. Precisely, the test accuracy is about 3\% higher for the entire face for 10 epochs. This is logical, as smile is not only visible on the mouth of humans, but also in other area, such as the cheeks. This margin reduces to about 2\% for 50 epochs.

Training time for the entire face is about 20\% higher for the entire face than for just the mouth. This is surprisingly less than expected since the number of input pixels is nearly double than for the mouth.

\section{Model selection for reduced dataset}
\label{chapter:red}
In the model selection in Chapter~\ref{chapter:full}, the entire DISFA database was used. As measured in Chapter~\ref{chapter:DISFA}, 48,612 of the entire 130,788 images are neutral, meaning no action units are set. The more neutral images, the easier the training of the smile detector.
Therefore, in this section, a reduced set of DISFA is used. It consists of all 82,176 images that have some action unit(s) sets and 30\% of the 48,612 remaining neutral images, making 96,759 images in total.

The setup of the experiments remains the same, in particular the ratio of training/validation/test examples, the two different input times, mouth and face, the number of epochs and the non-exhaustive model selection. Training time statistics of each trained permutation are available in Appendix~\ref{chapter:time}.

\subsection{Mouth}
For the mouth input, Tables~\ref{table:cv_mouth_complete_10_red} and \ref{table:cv_mouth_complete_50_red} contain the result of the model selection for 10 and 50 epochs, respectively.
For 10 epochs, for none of the four parameters the default one was chosen to be better than the ones available in the selection. In particular, lowest losses are returned for 2 convolutions, 2 hidden layers, 400 hidden units per layer and a dropout value of 0.
Overall, test accuracies are on a high level and individually tested values for the parameters have minor effect as test accuracies range from $89.46\%$ to $93.62\%$. 2 convolutions result in the highest accuracy, whereas a dropout value of $0.7$ reduces the test accuracy the most.

For 50 epochs, these selected values remain the same, except that in this case 300 hidden units have the lowest test loss.
Overall, test accuracies are on a even higher level and individually tested values for the parameters have minor effect as test accuracies range from $95.84\%$ to $97.59\%$. A dropout value of $0$ results in the highest accuracy, whereas all default values combined reduces the test accuracy the most.

\subsection{Face}

Conversely, for the entire face input, Tables~\ref{table:cv_face_complete_10_red} and \ref{table:cv_face_complete_50_red} contain the result of the model selection for 10 and 50 epochs, respectively.
For 10 epochs, the default one was chosen to be better than the ones available in the selection for the number of convolutions. In particular, lowest losses are returned for 1 convolution, 2 hidden layers, 300 hidden units per layer and a dropout value of 0.
Overall, test accuracies are on a high level and individually tested values for the parameters have minor effect as test accuracies range from $92.65\%$ to $95.44\%$. A dropout value of $0$ results in the highest accuracy, whereas 3 convolutions reduce the test accuracy the most.

For 50 epochs, these selected values remain the same for the number of convolutions and number of hidden units. The optimal number of hidden layers and dropout change to 1 and 0.1, respectively.
Overall, test accuracies are on a even higher level and individually tested values for the parameters have minor effect as test accuracies range from $95.91\%$ to $98.16\%$. A dropout value of $0.1$ results in the highest accuracy, whereas 3 hidden layers reduce the test accuracy the most.

\subsection{Comparison of mouth vs. face}
For both input parts, mouth and face, the model selection for 50 epochs returned different optimal parameters, which are collected in Table~\ref{table:cv_both_final_models_red}.

\begin{table}[h!]
\centering
\begin{tabular}{ c || c | c | c | c }
Input & \#Convs & \#Hidden layers & \#Units hidden layers & Dropout \\
\hline
\hline
Mouth & 2 & 2 & 300 & 0 \\
 \hline
Face & 1 & 1 & 300 & 0.1 \\
\end{tabular}
\caption{Selected parameter values for mouth and face input. For reduced dataset.}
\label{table:cv_both_final_models_red}
\end{table}

Overall, the test accuracies are slightly higher for the input of the entire face than just the mouth for 10 and 50 epochs. Precisely, the test accuracy is about 2\% higher for the entire face for 10 epochs. This margin reduces to about 1\% for 50 epochs.

Training time for the entire face is also about 20\% higher for the entire face than for just the mouth.

\section{Repeatability of experiments}
\label{chapter:repeat}
Each experiment was performed exactly once. Training of neural networks is subject to a random initialization of the weights at the beginning of the training and to the random split of the data into training, validation and test sets. Therefore, repeating an experiment may return different results.
If this difference is large, each experiment must be conducted for multiple times to use its median in the model selection decisions.
In order to assess if such a time-consuming process is necessary or not, the training of the neural network for 2 hidden layers for the mouth input in the model selection was conducted 10 times for the full dataset.
The results are available in Table~\ref{table:cv_10_times} with standard deviation of $0.041725\%$ in the test accuracy.
Because of this low standard deviation, performing each experiment exactly once has only a very low bias and is therefore relatively safe to do for reasons of faster training time.
The standard deviation of the cross-entropy loss has been omitted as it is not meaningful to humans.

\begin{table}[h!]
\centering
\begin{tabular}{ c || c }
Experiment number & Test accuracy \\
\hline
\hline
1 & 97.58\% \\
\hline
2 & 97.51\% \\
\hline
3 & 97.59\% \\
\hline
4 & 97.49\% \\
\hline
5 & 97.55\% \\
\hline
\textbf{6} & \textbf{97.62\%} \\
\hline
7 & 97.59\% \\
\hline
8 & 97.57\% \\
\hline
9 & 97.52\% \\
\hline
10 & 97.61\% \\
\hline
\hline
Standard deviation & 0.041725\% \\
\end{tabular}
\caption{Repeatability of training of architecture with default values and 2 hidden layers for mouth for 50 epochs: standard deviation of test accuracies. Optimal values in \textbf{bold}. For full dataset.}
\label{table:cv_10_times}
\end{table}

\section{Evaluation of final models for full and reduced datasets}
\label{chapter:evaluation_final}
In this section, the performance of the final models composed of the values selected in Chapters~\ref{chapter:full} and \ref{chapter:red} for the full and reduced datasets, respectively, are reported.

\begin{figure}[h!]
    \centering
    \includegraphics[width=0.7\textwidth]{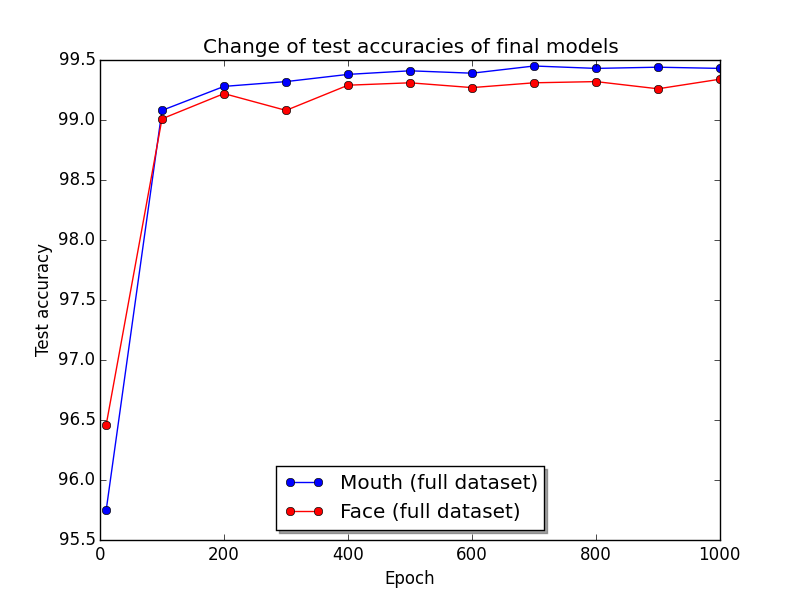}
    \caption{Change of test accuracy for mouth and face data over 1000 epochs. For full dataset.}
    \label{fig:1000_epochs_acc_full}
\end{figure}

For the full dataset, the final models selected in Table~\ref{table:cv_both_final_models_full} were trained for up to 1000 epochs. Table~\ref{table:cv_both_complete_full} contains a selection of test losses and accuracies of both models.
The best accuracies are 99.45\% and 99.34\% for the mouth and face input, respectively. The full results are available in Appendix~\ref{chapter:all_epochs}, for which the test accuracies are plotted in Figure~\ref{fig:1000_epochs_acc_full}.

\begin{table}[h!]
\centering
\begin{tabular}{ c || c | c || c | c}
 & \multicolumn{2}{c}{Mouth} & \multicolumn{2}{c}{Face} \\
\#Epochs & Test loss & Test accuracy & Test loss & Test accuracy \\
\hline
\hline
10 & 0.114402 & 95.75\% & 0.094356 & 96.46\% \\
 \hline
 100 & 0.027658 & 99.08\% & 0.030599 & 99.01\% \\
 \hline
 200 & \textbf{0.025298} & 99.28\% & \textbf{0.027087} & 99.22\% \\
 \hline
 700 & 0.033508 & \textbf{99.45\%} & 0.039649 & 99.31\% \\
 \hline
 1000 & 0.038099 & 99.43\% & 0.044800 & \textbf{99.34\%}
\end{tabular}
\caption{Result of model selection for mouth and face with the combined parameters for selected epochs. Optimal values per part in \textbf{bold}. For full dataset.}
\label{table:cv_both_complete_full}
\end{table}

For both inputs, the training is near to the best results after 200 epochs, after which the training wanders around the maximum. For the mouth and face input, the best accuracies are achieved after 700 and 1000 epochs, respectively. For the test loss however, the minima are achieved after 200 epochs.
This is a case in which accuracy and cross-entropy are not fully comparable.

\begin{figure}[h!]
    \centering
    \includegraphics[width=0.7\textwidth]{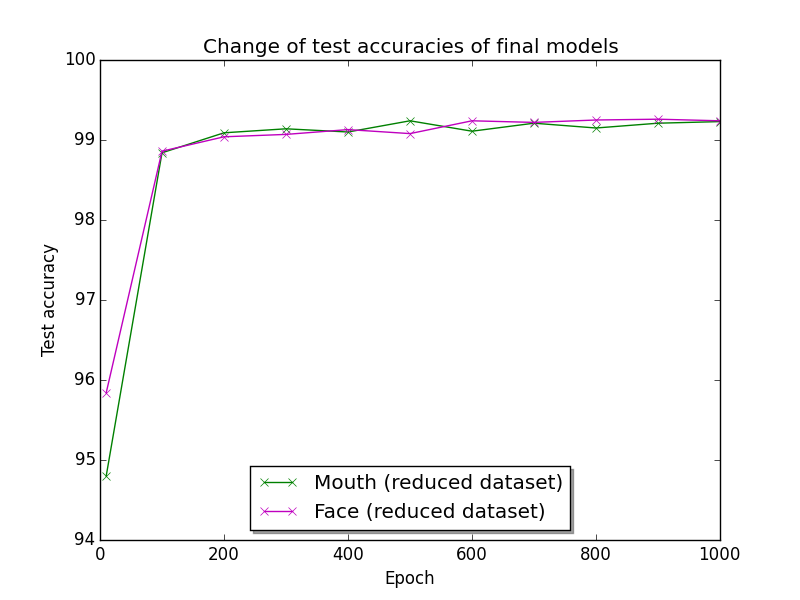}
    \caption{Change of test accuracy for mouth and face data over 1000 epochs. For reduced dataset.}
    \label{fig:1000_epochs_acc_red}
\end{figure}

For the reduced dataset, the final models selected in Table~\ref{table:cv_both_final_models_red} were also trained for up to 1000 epochs. Table~\ref{table:cv_both_complete_red} contains a selection of test losses and accuracies of both models.
The best accuracies are 99.24\% and 99.26\% for the mouth and face input, respectively. The full results are available in Appendix~\ref{chapter:all_epochs}, for which the test accuracies are plotted in Figure~\ref{fig:1000_epochs_acc_red}.

\begin{table}[h!]
\centering
\begin{tabular}{ c || c | c || c | c}
 & \multicolumn{2}{c}{Mouth} & \multicolumn{2}{c}{Face} \\
\#Epochs & Test loss & Test accuracy & Test loss & Test accuracy \\
\hline
\hline
10 & 0.134788 & 94.80\% & 0.109536 & 95.84\% \\
 \hline
100 & \textbf{0.036598} & 98.84\% & 0.033194 & 98.86\% \\
 \hline
 500 & 0.044365 & \textbf{99.24}\% & 0.031884 & 99.08\% \\
 \hline
 700 & 0.043212 & 99.21\% & \textbf{0.027191} & 99.22\% \\
 \hline
 900 & 0.042291 & 99.21\% & 0.027501 & \textbf{99.26\%} \\
 \hline
1000 & 0.041232 & 99.23\% & 0.030611 & 99.24\%
\end{tabular}
\caption{Result of model selection for mouth and face with the combined parameters for selected epochs. Optimal values per part in \textbf{bold}. For reduced dataset.}
\label{table:cv_both_complete_red}
\end{table}

For both inputs, the training is near to the best results after 200 epochs, after which the training wanders around the maximum. For the mouth and face input, the best accuracies are achieved after 500 and 900 epochs, respectively. For the test loss however, the minima are achieved after 100 and 700 epochs, respectively.
This is another case in which accuracy and cross-entropy are not fully comparable.

\begin{figure}[h!]
    \centering
    \includegraphics[width=0.7\textwidth]{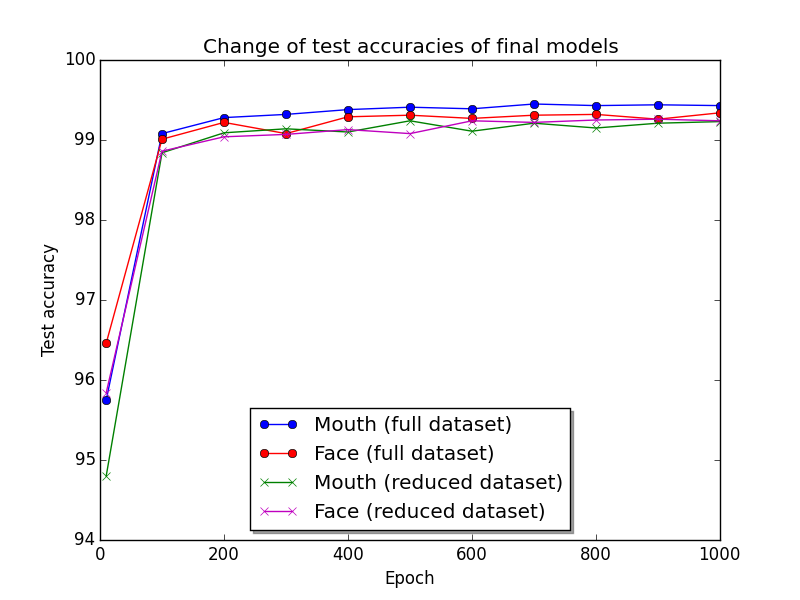}
    \caption{Change of test accuracy for mouth and face data over 1000 epochs. For both datasets.}
    \label{fig:1000_epochs_acc}
\end{figure}

Comparing the performance of the final models for both datasets, the models perform slightly better for the full dataset. In particular, for the mouth input, using the full dataset outperforms the reduced dataset with 99.45\% over 99.24\% for the mouth input. For the face input, the full dataset also outperforms the reduced dataset with 99.34\% over 99.26\%.
The comparison of those four accuracies is visualized in Figure~\ref{fig:1000_epochs_acc}.

Overall, using the full dataset, the models perform slightly better than using the reduced dataset. This makes intuitively sense, but the overall gap is little and may also be caused by the individual parameter values selected in the model selection. Nonetheless, this gap is much smaller than initially expected since 70\% of the neutral images have been dropped.
As discussed in Chapter~\ref{chapter:DISFA}, the original accuracies for smile in the DISFA database are 65.55\%, 72.94\% and 79.67\%, depending on the concrete feature description.
The best accuracy of 99.45\% was achieved in this project for the full DISFA dataset using the mouth input.
Overall, very high accuracies have been achieved in the experiments.
They are not fully comparable to the original DISFA paper in which a SVM was trained for a multi-class intensity problem.

\section{Comparison of low and high intensities for reduced dataset}
\label{chapter:low_high}
In this section, the experiment of Chapter~\ref{chapter:red} is repeated under different conditions. DISFA intensities range from 0-5, with 5 being the strongest intensity, for which Chapter~\ref{chapter:DISFA} contains the distribution of AU12.
In the following, intensities 1 and 2 are grouped together under the name low intensities, whereas intensities 4 and 5 are grouped together under the name high intensities.

For the low intensities, there are 72,194 images that have some action unit(s) set, and of those that have AU12 set, the intensities are 1 or 2. Furthermore, there are again 48,612 neutral ones. Similar to the reference experiment in Chapter~\ref{chapter:red}, 30\% of the 48,612 remaining neutral images are chosen, making 86,777 images in total.

Due to lack of time, no model selection could be performed. Instead, the parameter values chosen in Chapter~\ref{chapter:red} are used, since that experiment is the one most similar to this one. Overall, the exact parameter values have proven to be of less importance in the previous experiments for sufficiently many epochs, as summarized in Chapter~\ref{chapter:model_selection_res}.
Table~\ref{table:cv_both_final_models_red_low_high} contains the chosen parameter values for this experiment.

\begin{table}[h!]
\centering
\begin{tabular}{ c || c | c | c | c }
Input & \#Convs & \#Hidden layers & \#Units hidden layers & Dropout \\
\hline
\hline
Mouth & 2 & 2 & 300 & 0 \\
 \hline
Face & 1 & 1 & 300 & 0.1 \\
\end{tabular}
\caption{Parameter values for mouth and face input for low and high intensity models.}
\label{table:cv_both_final_models_red_low_high}
\end{table}

As measured in Chapter~\ref{chapter:evaluation_final}, only a few hundred epochs were necessary for the final models to get very close to the maximum accuracies. More epochs only had a minor effect, if at all, or may have even caused slight overfitting. Due to lack of time and based on these considerations, all models in this section are only trained for up to 400 epochs. Chapter~\ref{chapter:time} contains the training times per epoch of the respective models.

Table~\ref{table:cv_both_complete_red_low} contains the test losses and accuracies of the low intensity models, for mouth and face input, respectively. 
For the mouth input, the best test accuracy is achieved after 300 epochs with 98.96\%. Conversely, for the face input, the best test accuracy is achieved after 400 epochs with 99.08\%.
This difference of $0.12\%$ may be caused by various factors, including the lack of model selection, the number of epochs or general bias due to random initializations and random split of sets (see Chapter~\ref{chapter:repeat}).

\begin{table}[h!]
\centering
\begin{tabular}{ c || c | c || c | c}
 & \multicolumn{2}{c}{Mouth} & \multicolumn{2}{c}{Face} \\
\#Epochs & Test loss & Test accuracy & Test loss & Test accuracy \\
\hline
\hline
10 & 0.149945 & 93.85\% & 0.119456 & 95.20\% \\
 \hline
50 & 0.055449 & 98.28\% & 0.055006 & 97.85\% \\
 \hline
 100 & 0.057868 & 98.52\% & 0.039254 & 98.53\% \\
 \hline
 200 & \textbf{0.056766} & 98.79\% & 0.032467 & 98.94\% \\
 \hline
 300 & 0.064010 & \textbf{98.96\%} & 0.034236 & 98.93\% \\
 \hline
400 & 0.068849 & 98.95\% & \textbf{0.030127} & \textbf{99.08}\%
\end{tabular}
\caption{Result of training for mouth and face with the combined parameters for up to 400 epochs for low intensity models. Optimal values per part in \textbf{bold}.}
\label{table:cv_both_complete_red_low}
\end{table}

The same experiment is repeated for the high intensity models.
For the high intensities, there are 54,133 images that have some action unit(s) set, and of those that have AU12 set, the intensities are 4 or 5. Furthermore, there are again 48,612 neutral ones. Also, similar to the reference experiment in Chapter~\ref{chapter:red}, 30\% of the 48,612 remaining neutral images are chosen, making 68,716 images in total.
The same models as for the low intensities in Table~\ref{table:cv_both_final_models_red_low_high} are chosen and the experiments are run for 400 epochs each.

Table~\ref{table:cv_both_complete_red_high} contains the test losses and accuracies of the low intensity models, for mouth and face input, respectively. 
For the mouth input, the best test accuracy is achieved after 100 epochs with 99.94\%. After that, the test accuracy converges, but the test loss increases slightly, indicating the model to slightly overfit.
Conversely, for the face input, the best test accuracy is achieved after 200 epochs with 99.98\%.
This difference of $0.04\%$ may be also caused by various factors, including the lack of model selection, the number of epochs or general bias due to random initializations and random split of sets (see Chapter~\ref{chapter:repeat}).

\begin{table}[h!]
\centering
\begin{tabular}{ c || c | c || c | c}
 & \multicolumn{2}{c}{Mouth} & \multicolumn{2}{c}{Face} \\
\#Epochs & Test loss & Test accuracy & Test loss & Test accuracy \\
\hline
\hline
10 & 0.027446 & 99.17\% & 0.007468 & 99.85\% \\
 \hline
50 & 0.006950 & 99.93\% & 0.003457 & 99.94\% \\
 \hline
 100 & \textbf{0.009340} & \textbf{99.94\%} & 0.004088 & 99.96\% \\
 \hline
 200 & 0.011527 & 99.94\% & 0.003399 & \textbf{99.98\%} \\
 \hline
 300 & 0.012338 & 99.94\% & 0.003556 & 99.96\% \\
 \hline
400 & 0.012862 & 99.94\% & \textbf{0.003347} & 99.97\%
\end{tabular}
\caption{Result of training for mouth and face with the combined parameters for up to 400 epochs for high intensity models. Optimal values per part in \textbf{bold}.}
\label{table:cv_both_complete_red_high}
\end{table}

For both, low and high intensity smiles, the gap between accuracies for mouth and face input per intensity group are small.
Both models for the high intensity smiles perform nearly 1\% better than for the low intensity smiles.

\begin{figure}[h!]
    \centering
    \subfloat[No smile\label{subfig-1:dummy}]{%
        \includegraphics[width=0.3\textwidth]{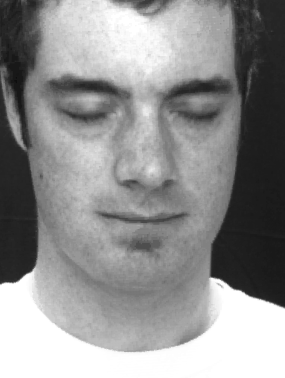}
    }%
    \hfill
    \subfloat[Low intensity smile\label{subfig-2:dummy}]{%
      \includegraphics[width=0.3\textwidth]{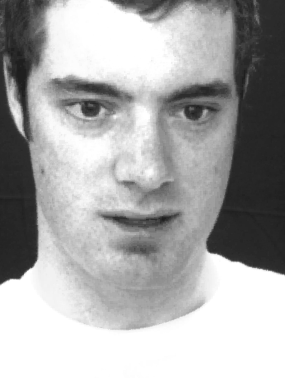}
    }%
    \hfill
    \subfloat[High intensity smile\label{subfig-2:dummy}]{%
      \includegraphics[width=0.3\textwidth]{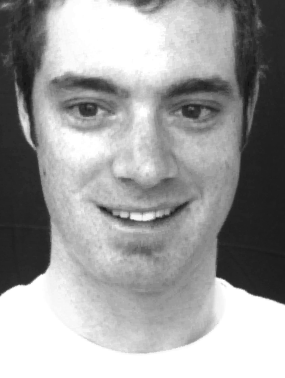}
    }%
    \caption{DISFA examples of video 002 for no smile, low intensity smile and high intensity smile \cite{DISFA}.}
    \label{fig:disfa_example_1}
\end{figure}

\begin{figure}[h!]
    \centering
    \subfloat[No smile\label{subfig-1:dummy}]{%
        \includegraphics[width=0.3\textwidth]{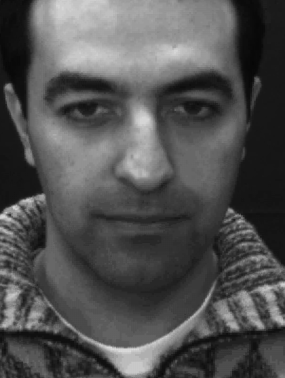}
    }%
    \hfill
    \subfloat[Low intensity smile\label{subfig-2:dummy}]{%
      \includegraphics[width=0.3\textwidth]{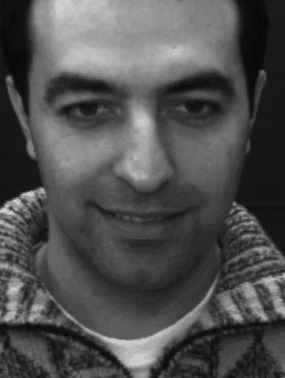}
    }%
    \hfill
    \subfloat[High intensity smile\label{subfig-2:dummy}]{%
      \includegraphics[width=0.3\textwidth]{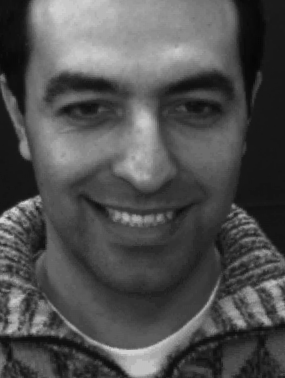}
    }%
    \caption{DISFA examples of video 005 for no smile, low intensity smile and high intensity smile \cite{DISFA}.}
    \label{fig:disfa_example_2}
\end{figure}

\begin{figure}[h!]
    \centering
    \subfloat[No smile\label{subfig-1:dummy}]{%
        \includegraphics[width=0.3\textwidth]{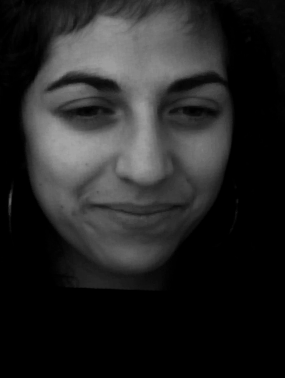}
    }%
    \hfill
    \subfloat[Low intensity smile\label{subfig-2:dummy}]{%
      \includegraphics[width=0.3\textwidth]{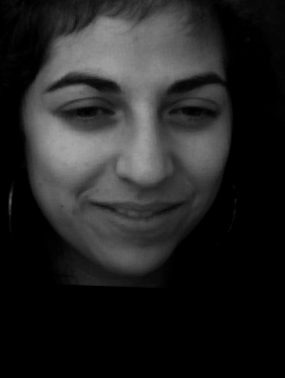}
    }%
    \hfill
    \subfloat[High intensity smile\label{subfig-2:dummy}]{%
      \includegraphics[width=0.3\textwidth]{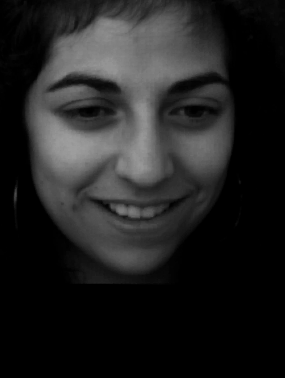}
    }%
    \caption{DISFA examples of video 023 for no smile, low intensity smile and high intensity smile \cite{DISFA}.}
    \label{fig:disfa_example_3}
\end{figure}

It is interesting to investigate this behavior further, since in Chapter~\ref{chapter:DISFA} it was measured that there are 20,810 low intensity smile images and only 2,749 high intensity smile images. 
Figures~\ref{fig:disfa_example_1}, \ref{fig:disfa_example_2} and \ref{fig:disfa_example_3} contain example images of three different videos for no smile, low intensity smile and high intensity smile. In the examples however, other action units may be set, too.

Comparing those examples, the size of the mouth changes just a bit for low intensity smiles, whereas the size, and in particular the height, change a lot for high intensity smiles. Also, the teeth are much more visible for high intensity smiles than for low intensity smiles.
Aside from the mouth, there are significant changes in the entire face, too: the muscles of the cheeks look totally different for smiles than for no smile, with a stronger change of the cheeks for high intensity smiles.

These factors contribute a lot to the high accuracies achieved in both experiments in different ways: for low intensity smiles, there is much more training data for the neural network in oder to discriminate between smile or no smile.
For high intensity smiles however, there is much less training data available, yet the changes in the mouth and around the cheeks are significant. Therefore, also for less training data in this experiment, very high accuracies can be achieved, even higher than for the low intensities.

\section{Classification of low and high intensities}
The previous experiments considered aside from AU12 set also other action units set in the training data.
The experiments and comparisons in Chapter~\ref{chapter:low_high} revealed interesting observations for the discrimination of high or low intensity smiles against the remaining (reduced) DISFA dataset.
In this section, only low and high intensity smiles are kept for discrimination.
Based on Chapter~\ref{chapter:DISFA}, there are 20,810 low intensity smile images and 2,749 high intensity smile images, making 23,559 images in total.
The same models from Table~\ref{table:cv_both_final_models_red_low_high} are chosen in this experiment, that were also trained for 400 epochs.

Table~\ref{table:cv_both_complete_high_vs_red} contains the results of this experiment. For the mouth input, an accuracy of 99.82\% is achieved after 200 epochs and then converges. The test loss slightly increases from then on, whereas the test accuracy remains the same.
For the face input, an accuracy of 99.87\% is achieved after 300 epochs.

For both inputs, very high accuracies are achieved. Due to the lack of model selection and the general bias in these experiments due to random initializations and random split in sets, it is difficult to say if the mouth only or the entire face input is beneficial.

\begin{table}[h!]
\centering
\begin{tabular}{ c || c | c || c | c}
 & \multicolumn{2}{c}{Mouth} & \multicolumn{2}{c}{Face} \\
\#Epochs & Test loss & Test accuracy & Test loss & Test accuracy \\
\hline
\hline
10 & 0.042481 & 98.78\% & 0.027992 & 99.22\% \\
 \hline
50 & 0.008517 & 99.80\% & 0.008929 & 99.69\% \\
 \hline
100 & 0.010142 & 99.73\% & 0.005758 & 99.80\% \\
 \hline
200 & \textbf{0.008854} & \textbf{99.82\%} & 0.004467 & 99.76\% \\
 \hline
300 & 0.009222 & 99.82\% & \textbf{0.002678} & \textbf{99.87\%} \\
 \hline
400 & 0.009713 & 99.82\% & 0.003238 & 99.84\% \\
\end{tabular}
\caption{Result of training for mouth and face with the combined parameters for up to 400 epochs for classification of low and high intensities. Optimal values per part in \textbf{bold}.}
\label{table:cv_both_complete_high_vs_red}
\end{table}

\chapter{Conclusions and future work}
\label{chapter:conclusions}
Deep learning is a set of powerful machine learning algorithms and concepts with groundbreaking success for the last ten years. The main benefit of deep neural networks are their ability to learn complex non-linear hypothesis without the need of explicitly modeling features, but rather learning them from data.
Convolutional networks allow to handle distortions, such as translation and rotation in the input, which occurs frequently in computer vision.
Applied to action unit recognition and smile recognition in particular, a deep convolutional neural network model with an overall accuracy of 99.45\% significantly outperforms existing approaches with accuracies ranging from 65.55\% to 79.67\%. The network parameter values are subject to extensive model selection.
Various variations of this experiment are run, such as retaining less neutral images or only high or low intensities or classifying into low or high intensities. For all experiments, very high accuracies above 98.90\% are achieved, too.
Choosing the entire face as input or just the mouth only led to minor differences in the accuracies, not generally favoring either input.

The proposed temporal part using LSTMs was not implemented due to the high accuracies achieved. It would however be interesting to implement it in the future in order to predict smiles in image sequences.
There are many further topics worth to be investigated. For example, instead of one CNN being trained on the entire face or the mouth, multiple CNNs could be trained on different regions of the face. Possible regions are the mouth, the nose and both eyes. A specialized CNN could be trained for each region allowing to generalize better because of the lower number of parameters in each network. The CNNs can then be combined using a Shape Boltzmann Machine \cite{shape_RBM}. Furthermore, in order to understand the networks better and to do informedly better than blind model selection, it would be interesting to visualize the units of a network to understand what they learned to detect. This can help to work well on a variety of action units other than smile.
In this thesis, the mouth was compared to the entire face, for which no significant difference was detected. It would however, be interesting to investigate this more by comparing the mouth to the face excluding the mouth.
So far, the existing model has been used for classification. In a next step, it can be adopted to regression of action unit intensities or even valence-arousal \cite{valence_arousal}.
To date, stochastic gradient descent is the preferred training algorithm for neural networks, as discovered by LeCun in the 1980s and 1990s \cite{lecun_research}. Overall, stochastic gradient descent performs well on deep neural networks, yet it would be interesting to investigate if LeCun's observation still holds for deep neural networks. Therefore, stochastic gradient descent should be compared to a variety of other optimization algorithms, such as Gaussian Newton or Quasi Newton methods when training deep neural networks.

\appendix
\chapter{Statistics of all action units}
\label{chapter:fullstats}
This appendix contains in Table~\ref{table:all_stats} the full statistics of action units of the databases considered in Chapter~\ref{chapter:databases}: AMFED, CASME, DISFA, GEMEP, MAHNOB Laughter and shoulder pain.

{\centering
\begin{longtable}{ c || c | c | c | c | c | c }
  & AMFED & CASME & DISFA & GEMEP & MAHNOB Laughter & Shoulder Pain \\
\hline
\hline
AU1 & - & 1976 & 8778 & 1584 & - & - \\
\hline
AU2 & 8500 & 936 & 7364 & 1618 & - & - \\
\hline
AU3 & - & - & - & 0 & - & - \\
\hline
AU4 & 9078 & 1937 & 24595 & 1342 & - & 1074 \\
\hline
AU5 & 5478 & - & 2729 & 735 & - & - \\
\hline
AU6 & - & 304 & 19484 & 1780 & - & 5557 \\
\hline
AU7 & - & 405 & - & 2100 & - & 3364 \\
\hline
AU8 & - & - & - & 7 & - & - \\
\hline
AU9 & 76 & 731 & 7132 & 392 & - & 423 \\
\hline
AU10 & 657 & 112 & - & 2008 & - & 525 \\
\hline
AU11 & - & - & - & 512 & - & - \\
\hline
AU12 & - & 264 & 30794 & 2692 & - & 6887 \\
\hline
AU13 & - & 39 & - & 63 & - & - \\
\hline
AU14 & 8529 & 493 & - & 213 & - & - \\
\hline
AU15 & 637 & 298 & 7862 & 1014 & - & - \\
\hline
AU16 & - & 126 & - & 310 & - & - \\
\hline
AU17 & 5482 & 293 & 12930 & 820 & - & - \\
\hline
AU18 & 1800 & 197 & - & 417 & - & - \\
\hline
AU19 & - & - & - & 90 & - & - \\
\hline
AU20 & - & 62 & 4532 & 480 & - & 706 \\
\hline
AU21 & - & - & - & 95 & - & - \\
\hline
AU22 & - & - & - & 169 & - & - \\
\hline
AU23 & - & 21 & - & 163 & - & - \\
\hline
AU24 & - & 112 & - & 124 & - & - \\
\hline
AU25 & - & 132 & 46052 & 874 & - & 2406 \\
\hline
AU26 & 9626 & 165 & 24976 & 544 & - & 2092 \\
\hline
AU27 & - & - & - & 27 & - & 18 \\
\hline
AU28 & - & - & - & 39 & - & - \\
\hline
AU29 & - & - & - & 0 & - & - \\
\hline
AU30 & - & 19 & - & 197 & - & - \\
\hline
AU31 & - & - & - & 13 & - & - \\
\hline
AU32 & - & - & - & 24 & - & - \\
\hline
AU33 & - & - & - & 0 & - & - \\
\hline
AU34 & - & 38 & - & 6 & - & - \\
\hline
AU35 & - & - & - & 0 & - & - \\
\hline
AU36 & - & - & - & 5 & - & - \\
\hline
AU37 & - & - & - & 0 & - & - \\
\hline
AU38 & - & - & - & 46 & - & - \\
\hline
AU39 & - & - & - & 9 & - & - \\
\hline
AU40 & - & - & - & 0 & - & - \\
\hline
AU41 & - & - & - & 0 & - & - \\
\hline
AU42 & - & - & - & 0 & - & - \\
\hline
AU43 & - & - & - & 539 & - & 2434 \\
\hline
AU44 & - & - & - & 0 & - & - \\
\hline
AU45 & - & - & - & 167 & - & - \\
\hline
AU46 & - & - & - & 0 & - & - \\
\hline
AU47 & - & - & - & 0 & - & - \\
\hline
AU48 & - & - & - & 0 & - & - \\
\hline
AU49 & - & - & - & 0 & - & - \\
\hline
AU50 & - & - & - & 3929 & - & - \\
\hline
AU57 & 0 & - & - & - & - & - \\
\hline
AU58 & 0 & - & - & - & - & - \\
\hline
L1 & - & 223 & - & - & - & - \\
\hline
L2 & - & 52 & - & - & - & - \\
\hline
L9 & - & 25 & - & - & - & - \\
\hline
L10 & - & 46 & - & - & - & - \\
\hline
L12 & - & 85 & - & - & - & - \\
\hline
L14 & - & 45 & - & - & - & - \\
\hline
L15 & - & 16 & - & - & - & - \\
\hline
L20 & - & 31 & - & - & - & - \\
\hline
Laughter & - & - & - & - & 6404 & - \\
\hline
R1 & - & 88 & - & - & - & - \\
\hline
R2 & - & 90 & - & - & - & - \\
\hline
R9 & - & 13 & - & - & - & - \\
\hline
R10 & - & 21 & - & - & - & - \\
\hline
R12 & - & 0 & - & - & - & - \\
\hline
R14 & - & 18 & - & - & - & - \\
\hline
Smile & 77062 & - & - & - & - & - \\
\hline
negAU12 & 350 & - & - & - & - & - \\
\caption{Complete statistics of action units in databases: an integer denotes the number of frames in which an action unit is set (intensity $> 0$). A hyphen indicates that an action unit is not available in a database.}
\label{table:all_stats}
\end{longtable}}

\chapter{Training time of networks}
\label{chapter:time}
This appendix contains the full statistics of median epoch durations of the different convolutional models trained in Chapter~\ref{chapter:resevaluation}. For none of the experiments, a significant spread of the epoch durations was observed, therefore only the median is available in this chapter.

\section{Full dataset}
Table~\ref{table:cv_epoch_time1_full} shows that training time mostly depends on the number of convolutional and pooling layers. Different levels of dropout have no significant impact on the duration of an epoch. Training time is slightly less when dropout is disabled totally for $p = 0$.
The number of hidden units in the tested range has only a minor effect on the training time, but grows slowly with the increased values. Significantly larger number of units per layer are likely to change the training time stronger.

\begin{table}[h!]
\centering
\begin{tabular}{ c | c | c | c || c | c }
\#Convs & \#Hidden layers & \#Units hidden layers & Dropout & Mouth & Face \\
\hline
\hline
1 & 1 & 100 & 0.5 & 46.889 & 52.367 \\
 \hline
  \hline
2 & 1 & 100 & 0.5 & 110.956 & 123.221 \\
 \hline
3 & 1 & 100 & 0.5 & 135.143 & 150.961 \\
 \hline
  \hline
1 & 2 & 100 & 0.5 & 46.844 & 52.382 \\
 \hline
1 & 3 & 100 & 0.5 & 47.029 & 51.993  \\
 \hline
  \hline
1 & 1 & 200 & 0.5 & 49.069 & 54.663 \\
 \hline
1 & 1 & 300 & 0.5 & 50.632 & 56.578 \\
 \hline
1 & 1 & 400 & 0.5 & 54.916 & 61.095 \\
 \hline
  \hline
1 & 1 & 100 & 0 & 45.034 & 49.843 \\
 \hline
1 & 1 & 100 & 0.1 & 47.353 & 52.297 \\
 \hline
1 & 1 & 100 & 0.7 & 47.107 & 52.247 \\
\end{tabular}
\caption{Median epoch duration in seconds during model selection of different architectures. For full dataset.}
\label{table:cv_epoch_time1_full}
\end{table}

Table~\ref{table:cv_epoch_time2_full} contains the median epoch duration for the final models selected for mouth and face input. A comparison of the training time is not possible, as the selected values are very different and because the number of input pixels is different, too.

\begin{table}[h!]
\centering
\begin{tabular}{ c | c | c | c || c | c }
\#Convs & \#Hidden layers & \#Units hidden layers & Dropout & Mouth & Face \\
\hline
\hline
2 & 2 & 400 & 0.1 & 112.436 & - \\
 \hline
1 & 1 & 400 & 0 & - & 58.823 \\
\end{tabular}
\caption{Median epoch duration in seconds for final models selected. For full dataset.}
\label{table:cv_epoch_time2_full}
\end{table}

\section{Reduced dataset}
Table~\ref{table:cv_epoch_time1_red} shows that training time also mostly depends on the number of convolutional and pooling layers. The same observations as for the full dataset apply to the reduced dataset.

\begin{table}[h!]
\centering
\begin{tabular}{ c | c | c | c || c | c }
\#Convs & \#Hidden layers & \#Units hidden layers & Dropout & Mouth & Face \\
\hline
\hline
1 & 1 & 100 & 0.5 & 35.137 & 38.807 \\
 \hline
  \hline
2 & 1 & 100 & 0.5 & 82.415 & 91.555 \\
 \hline
3 & 1 & 100 & 0.5 & 100.414 & 112.321 \\
 \hline
  \hline
1 & 2 & 100 & 0.5 & 35.301 & 38.920 \\
 \hline
1 & 3 & 100 & 0.5 & 35.564 & 38.995  \\
 \hline
  \hline
1 & 1 & 200 & 0.5 & 36.480 & 40.851 \\
 \hline
1 & 1 & 300 & 0.5 & 37.645 & 41.751 \\
 \hline
1 & 1 & 400 & 0.5 & 41.113 & 45.364 \\
 \hline
  \hline
1 & 1 & 100 & 0 & 33.621 & 37.118 \\
 \hline
1 & 1 & 100 & 0.1 & 34.934 & 38.977 \\
 \hline
1 & 1 & 100 & 0.7 & 34.870 & 38.715 \\
\end{tabular}
\caption{Median epoch duration in seconds during model selection of different architectures. For reduced dataset.}
\label{table:cv_epoch_time1_red}
\end{table}
 
Table~\ref{table:cv_epoch_time2_red} contains the median epoch duration for the final models selected for mouth and face input.

\begin{table}[h!]
\centering
\begin{tabular}{ c | c | c | c || c | c }
\#Convs & \#Hidden layers & \#Units hidden layers & Dropout & Mouth & Face \\
\hline
\hline
2 & 2 & 300 & 0 & 82.668 & - \\
 \hline
1 & 1 & 300 & 0.1 & - & 41.485 \\
\end{tabular}
\caption{Median epoch duration in seconds for final models selected. For reduced dataset.}
\label{table:cv_epoch_time2_red}
\end{table}

\section{Low and high intensities for reduced dataset}
Tables~\ref{table:cv_epoch_time_red_low} and \ref{table:cv_epoch_time_red_high} contain the median epoch duration for the models for mouth and face input for low and high intensities, respectively.

\begin{table}[h!]
\centering
\begin{tabular}{ c | c | c | c || c | c }
\#Convs & \#Hidden layers & \#Units hidden layers & Dropout & Mouth & Face \\
\hline
\hline
2 & 2 & 300 & 0 & 73.989 & - \\
 \hline
1 & 1 & 300 & 0.1 & - & 37.491 \\
\end{tabular}
\caption{Median epoch duration in seconds for models for low intensities. For reduced dataset.}
\label{table:cv_epoch_time_red_low}
\end{table}

\begin{table}[h!]
\centering
\begin{tabular}{ c | c | c | c || c | c }
\#Convs & \#Hidden layers & \#Units hidden layers & Dropout & Mouth & Face \\
\hline
\hline
2 & 2 & 300 & 0 & 58.094 & - \\
 \hline
1 & 1 & 300 & 0.1 & - & 29.625 \\
\end{tabular}
\caption{Median epoch duration in seconds for models for high intensities. For reduced dataset.}
\label{table:cv_epoch_time_red_high}
\end{table}

\section{Classification of low and high intensities}
Table~\ref{table:cv_epoch_time_low_vs_high} contains the median epoch duration for the models for mouth and face input. 

\begin{table}[h!]
\centering
\begin{tabular}{ c | c | c | c || c | c }
\#Convs & \#Hidden layers & \#Units hidden layers & Dropout & Mouth & Face \\
\hline
\hline
2 & 2 & 300 & 0 & 19.865 & - \\
 \hline
1 & 1 & 300 & 0.1 & - & 10.071 \\
\end{tabular}
\caption{Median epoch duration in seconds for models for classification of low and high intensities.}
\label{table:cv_epoch_time_low_vs_high}
\end{table}

\chapter{Result of model selection}
\label{chapter:model_selection_res}
This appendix contains the full results of the model selection of the different convolutional models trained in Chapter~\ref{chapter:resevaluation}.

\section{Full dataset}
Tables~\ref{table:cv_mouth_complete_10_full} and \ref{table:cv_mouth_complete_50_full} contain the results of the model selection for the mouth input for 10 and 50 epochs, respectively.

\begin{table}[h!]
\centering
\begin{tabular}{ c | c | c | c || c | c }
\#Convs & \#Hidden layers & \#Units hidden layers & Dropout & Test loss & Test accuracy \\
\hline
\hline
1 & 1 & 100 & 0.5 & 0.171199 & 92.88\% \\
 \hline
  \hline
\textbf{2} & 1 & 100 & 0.5 & \textbf{0.148397} & \textbf{94.02\%} \\
 \hline
3 & 1 & 100 & 0.5 & 0.161987 & 93.62\% \\
 \hline
  \hline
1 & 2 & 100 & 0.5 & 0.165286 & 93.35\% \\
 \hline
1 & \textbf{3} & 100 & 0.5 & \textbf{0.149762} & \textbf{94.21\%}  \\
 \hline
  \hline
1 & 1 & 200 & 0.5 & 0.157015 & 93.64\% \\
 \hline
1 & 1 & 300 & 0.5 & 0.158230 & 93.77\% \\
 \hline
1 & 1 & \textbf{400} & 0.5 & \textbf{0.155952} & \textbf{93.83\%} \\
 \hline
  \hline
1 & 1 & 100 & 0 & 0.144069 & 94.42\% \\
 \hline
1 & 1 & 100 & \textbf{0.1} & \textbf{0.139177} & \textbf{94.54\%} \\
 \hline
1 & 1 & 100 & 0.7 & 0.194659 & 90.99\% \\
\end{tabular}
\caption{Model selection for mouth for 10 epochs. Optimal values per parameter in \textbf{bold}. For full dataset.}
\label{table:cv_mouth_complete_10_full}
\end{table}

\begin{table}[h!]
\centering
\begin{tabular}{ c | c | c | c || c | c }
\#Convs & \#Hidden layers & \#Units hidden layers & Dropout & Test loss & Test accuracy \\
\hline
\hline
1 & 1 & 100 & 0.5 & 0.073235 & 97.15\% \\
 \hline
  \hline
\textbf{2} & 1 & 100 & 0.5 & \textbf{0.063466} & \textbf{97.64\%} \\
 \hline
3 & 1 & 100 & 0.5 & 0.084795 & 96.90\% \\
 \hline
  \hline
1 & \textbf{2} & 100 & 0.5 & \textbf{0.064558} & \textbf{97.58\%} \\
 \hline
1 & 3 & 100 & 0.5 & 0.083804 & 96.60\% \\
 \hline
  \hline
1 & 1 & 200 & 0.5 & 0.073072 & 97.31\% \\
 \hline
1 & 1 & 300 & 0.5 & 0.078304 & 97.11\% \\
 \hline
1 & 1 & \textbf{400} & 0.5 & \textbf{0.069398} & \textbf{97.50\%} \\
 \hline
  \hline
1 & 1 & 100 & 0 & 0.139177 & 94.54\% \\
 \hline
1 & 1 & 100 & \textbf{0.1} & \textbf{0.060566} & \textbf{97.70\%} \\
 \hline
1 & 1 & 100 & 0.7 & 0.090101 & 96.56\% \\
\end{tabular}
\caption{Model selection for mouth for 50 epochs. Optimal values per parameter in \textbf{bold}. For full dataset.}
\label{table:cv_mouth_complete_50_full}
\end{table}

Tables~\ref{table:cv_face_complete_10_full} and \ref{table:cv_face_complete_50_full} contain the results of the model selection for the face input for 10 and 50 epochs, respectively.

\begin{table}[h!]
\centering
\begin{tabular}{ c | c | c | c || c | c }
\#Convs & \#Hidden layers & \#Units hidden layers & Dropout & Test loss & Test accuracy \\
\hline
\hline
\textbf{1} & 1 & \textbf{100} & 0.5 & \textbf{0.106219} & \textbf{96.13\%} \\
 \hline
  \hline
2 & 1 & 100 & 0.5 & 0.123406 & 95.00\% \\
 \hline
3 & 1 & 100 & 0.5 & 0.136121 & 94.69\% \\
 \hline
  \hline
1 & \textbf{2} & 100 & 0.5 & \textbf{0.097022} & \textbf{96.26\%} \\
 \hline
1 & 3 & 100 & 0.5 & 0.105416 & 96.18\% \\
 \hline
  \hline
1 & 1 & 200 & 0.5 & 0.106186 & 95.75\% \\
 \hline
1 & 1 & 300 & 0.5 & 0.102369 & 96.07\% \\
 \hline
1 & 1 & 400 & 0.5 & 0.102993 & 95.99\% \\
 \hline
  \hline
1 & 1 & 100 & \textbf{0} & \textbf{0.094154} & \textbf{96.36\%} \\
 \hline
1 & 1 & 100 & 0.1 & 0.101364 & 96.03\% \\
 \hline
1 & 1 & 100 & 0.7 & 0.125406 & 94.76\% \\
\end{tabular}
\caption{Model selection for face for 10 epochs. Optimal values per parameter in \textbf{bold}. For full dataset.}
\label{table:cv_face_complete_10_full}
\end{table}

\begin{table}[h!]
\centering
\begin{tabular}{ c | c | c | c || c | c }
\#Convs & \#Hidden layers & \#Units hidden layers & Dropout & Test loss & Test accuracy \\
\hline
\hline
\textbf{1} & \textbf{1} & 100 & 0.5 & \textbf{0.053476} & \textbf{98.02\%} \\
 \hline
  \hline
2 & 1 & 100 & 0.5 & 0.068358 & 97.50\% \\
 \hline
3 & 1 & 100 & 0.5 & 0.069488 & 97.39\% \\
 \hline
  \hline
1 & 2 & 100 & 0.5 & 0.055521 & 98.00\% \\
 \hline
1 & 3 & 100 & 0.5 & 0.064505 & 97.70\% \\
 \hline
  \hline
1 & 1 & 200 & 0.5 & 0.052052 & 98.02\% \\
 \hline
1 & 1 & 300 & 0.5 & 0.051262 & 98.10\% \\
 \hline
1 & 1 & \textbf{400} & 0.5 & \textbf{0.050844} & \textbf{98.13\%} \\
 \hline
  \hline
1 & 1 & 100 & \textbf{0} & \textbf{0.042898} & \textbf{98.57\%} \\
 \hline
1 & 1 & 100 & 0.1 & 0.043234 & 98.30\% \\
 \hline
1 & 1 & 100 & 0.7 & 0.071993 & 97.37\% \\
\end{tabular}
\caption{Model selection for face for 50 epochs. Optimal values per parameter in \textbf{bold}. For full dataset.}
\label{table:cv_face_complete_50_full}
\end{table}

\section{Reduced dataset}
Tables~\ref{table:cv_mouth_complete_10_red} and \ref{table:cv_mouth_complete_50_red} contain the results of the model selection for the mouth input for 10 and 50 epochs, respectively.

\begin{table}[h!]
\centering
\begin{tabular}{ c | c | c | c || c | c }
\#Convs & \#Hidden layers & \#Units hidden layers & Dropout & Test loss & Test accuracy \\
\hline
\hline
1 & 1 & 100 & 0.5 & 0.200161 & 91.84\% \\
 \hline
  \hline
\textbf{2} & 1 & 100 & 0.5 & \textbf{0.166920} & \textbf{93.62\%} \\
 \hline
3 & 1 & 100 & 0.5 & 0.194500 & 92.80\% \\
 \hline
  \hline
1 & \textbf{2} & 100 & 0.5 & \textbf{0.205978} & \textbf{92.05\%} \\
 \hline
1 & 3 & 100 & 0.5 & 0.221186 & 91.19\%  \\
 \hline
  \hline
1 & 1 & 200 & 0.5 & 0.204515 & 91.79\% \\
 \hline
1 & 1 & 300 & 0.5 & 0.202924 & 91.52\% \\
 \hline
1 & 1 & \textbf{400} & 0.5 & \textbf{0.200398} & \textbf{92.24\%} \\
 \hline
  \hline
1 & 1 & 100 & \textbf{0} & \textbf{0.191992} & \textbf{92.32\%} \\
 \hline
1 & 1 & 100 & 0.1 & 0.208275 & 91.45\% \\
 \hline
1 & 1 & 100 & 0.7 & 0.235162 & 89.46\% \\
\end{tabular}
\caption{Model selection for mouth for 10 epochs. Optimal values per parameter in \textbf{bold}. For reduced dataset.}
\label{table:cv_mouth_complete_10_red}
\end{table}

\begin{table}[h!]
\centering
\begin{tabular}{ c | c | c | c || c | c }
\#Convs & \#Hidden layers & \#Units hidden layers & Dropout & Test loss & Test accuracy \\
\hline
\hline
1 & 1 & 100 & 0.5 & 0.106992 & 95.84\% \\
 \hline
  \hline
\textbf{2} & 1 & 100 & 0.5 & \textbf{0.072185} & \textbf{97.57\%} \\
 \hline
3 & 1 & 100 & 0.5 & 0.095411 & 96.64\% \\
 \hline
  \hline
1 & \textbf{2} & 100 & 0.5 & \textbf{0.082234} & \textbf{96.91\%} \\
 \hline
1 & 3 & 100 & 0.5 & 0.083829 & 96.74\%  \\
 \hline
  \hline
1 & 1 & 200 & 0.5 & 0.093198 & 96.65\% \\
 \hline
1 & 1 & \textbf{300} & 0.5 & \textbf{0.082962} & \textbf{96.98\%} \\
 \hline
1 & 1 & 400 & 0.5 & 0.087152 & 96.77\% \\
 \hline
  \hline
1 & 1 & 100 & \textbf{0} & \textbf{0.066274} & \textbf{97.59\%} \\
 \hline
1 & 1 & 100 & 0.1 & 0.081724 & 96.85\% \\
 \hline
1 & 1 & 100 & 0.7 & 0.102717 & 96.21\% \\
\end{tabular}
\caption{Model selection for mouth for 50 epochs. Optimal values per parameter in \textbf{bold}. For reduced dataset.}
\label{table:cv_mouth_complete_50_red}
\end{table}

Tables~\ref{table:cv_face_complete_10_red} and \ref{table:cv_face_complete_50_red} contain the results of the model selection for the face input for 10 and 50 epochs, respectively.

\begin{table}[h!]
\centering
\begin{tabular}{ c | c | c | c || c | c }
\#Convs & \#Hidden layers & \#Units hidden layers & Dropout & Test loss & Test accuracy \\
\hline
\hline
\textbf{1} & 1 & 100 & 0.5 & \textbf{0.137569} & \textbf{94.47\%} \\
 \hline
  \hline
2 & 1 & 100 & 0.5 & 0.162332 & 93.71\% \\
 \hline
3 & 1 & 100 & 0.5 & 0.178035 & 92.65\% \\
 \hline
  \hline
1 & \textbf{2} & 100 & 0.5 & \textbf{0.123239} & \textbf{95.11\%} \\
 \hline
1 & 3 & 100 & 0.5 & 0.151889 & 94.48\%  \\
 \hline
  \hline
1 & 1 & 200 & 0.5 & 0.127802 & 94.91\% \\
 \hline
1 & 1 & \textbf{300} & 0.5 & \textbf{0.127247} & \textbf{95.08\%} \\
 \hline
1 & 1 & 400 & 0.5 & 0.130562 & 94.89\% \\
 \hline
  \hline
1 & 1 & 100 & \textbf{0} & \textbf{0.114597} & \textbf{95.44\%} \\
 \hline
1 & 1 & 100 & 0.1 & 0.118390 & 95.44\% \\
 \hline
1 & 1 & 100 & 0.7 & 0.153938 & 93.76\% \\
\end{tabular}
\caption{Model selection for face for 10 epochs. Optimal values per parameter in \textbf{bold}. For reduced dataset.}
\label{table:cv_face_complete_10_red}
\end{table}

\begin{table}[h!]
\centering
\begin{tabular}{ c | c | c | c || c | c }
\#Convs & \#Hidden layers & \#Units hidden layers & Dropout & Test loss & Test accuracy \\
\hline
\hline
\textbf{1} & \textbf{1} & 100 & 0.5 & \textbf{0.067837} & \textbf{97.44\%} \\
 \hline
  \hline
2 & 1 & 100 & 0.5 & 0.082837 & 96.77\% \\
 \hline
3 & 1 & 100 & 0.5 & 0.090266 & 96.64\% \\
 \hline
  \hline
1 & 2 & 100 & 0.5 & 0.069571 & 97.31\% \\
 \hline
1 & 3 & 100 & 0.5 & 0.099563 & 95.91\%  \\
 \hline
  \hline
1 & 1 & 200 & 0.5 & 0.065923 & 97.41\% \\
 \hline
1 & 1 & \textbf{300} & 0.5 & \textbf{0.062300} & \textbf{97.58\%} \\
 \hline
1 & 1 & 400 & 0.5 & 0.062894 & 97.54\% \\
 \hline
  \hline
1 & 1 & 100 & 0 & 0.083884 & 96.99\% \\
 \hline
1 & 1 & 100 & \textbf{0.1} & \textbf{0.049972} & \textbf{98.16\%} \\
 \hline
1 & 1 & 100 & 0.7 & 0.091003 & 96.62\% \\
\end{tabular}
\caption{Model selection for face for 50 epochs. Optimal values per parameter in \textbf{bold}. For reduced dataset.}
\label{table:cv_face_complete_50_red}
\end{table}

\chapter{Performance of selected models}
\label{chapter:all_epochs}
This appendix contains the performance of the final models based on the selected values in Chapter~\ref{chapter:resevaluation}.

\section{Full dataset}
Table~\ref{table:cv_both_complete_1000_full} contains the full full collection of test losses and test accuracies for the two selected models trained for inputs of mouth and face data.

\begin{table}[h!]
\centering
\begin{tabular}{ c || c | c || c | c}
 & \multicolumn{2}{c}{Mouth} & \multicolumn{2}{c}{Face} \\
\#Epochs & Test loss & Test accuracy & Test loss & Test accuracy \\
\hline
\hline
10 & 0.114402 & 95.75\% & 0.094356 & 96.46\% \\
 \hline
 100 & 0.027658 & 99.08\% & 0.030599 & 99.01\% \\
 \hline
 200 & \textbf{0.025298} & 99.28\% & \textbf{0.027087} & 99.22\% \\
 \hline
 300 & 0.030369 & 99.32\% & 0.033196 & 99.08\% \\
 \hline
 400 & 0.029371 & 99.38\% & 0.030376 & 99.29\% \\
 \hline
 500 & 0.031548 & 99.41\% & 0.034192 & 99.31\% \\
 \hline
 600 & 0.037023 & 99.39\% & 0.033860 & 99.27\% \\
 \hline
 700 & 0.033508 & \textbf{99.45\%} & 0.039649 & 99.31\% \\
 \hline
 800 & 0.036150 & 99.43\% & 0.040020 & 99.32\% \\
 \hline
 900 & 0.038760 & 99.44\% & 0.042119 & 99.26\% \\
 \hline
 1000 & 0.038099 & 99.43\% & 0.044800 & \textbf{99.34\%}
\end{tabular}
\caption{Result of model selection for mouth and face with the combined parameters for 10, 100, 200, ..., 1000 epochs. Optimal values per part in \textbf{bold}. For full dataset.}
\label{table:cv_both_complete_1000_full}
\end{table}

\section{Reduced dataset}
Table~\ref{table:cv_both_complete_1000_red} contains the full full collection of test losses and test accuracies for the two selected models trained for inputs of mouth and face data.

\begin{table}[h!]
\centering
\begin{tabular}{ c || c | c || c | c}
 & \multicolumn{2}{c}{Mouth} & \multicolumn{2}{c}{Face} \\
\#Epochs & Test loss & Test accuracy & Test loss & Test accuracy \\
\hline
\hline
10 & 0.134788 & 94.80\% & 0.109536 & 95.84\% \\
 \hline
100 & \textbf{0.036598} & 98.84\% & 0.033194 & 98.86\% \\
 \hline
 200 & 0.036625 & 99.09\% & 0.030154 & 99.04\% \\
 \hline
 300 & 0.042087 & 99.14\% & 0.027567 & 99.07\% \\
 \hline
 400 & 0.049574 & 99.10\% & 0.026899 & 99.13\% \\
 \hline
 500 & 0.044365 & \textbf{99.24}\% & 0.031884 & 99.08\% \\
 \hline
 600 & 0.052758 & 99.11\% & 0.028884 & 99.24\% \\
 \hline
 700 & 0.043212 & 99.21\% & \textbf{0.027191} & 99.22\% \\
 \hline
 800 & 0.044299 & 99.15\% & 0.028185 & 99.25\% \\
 \hline
 900 & 0.042291 & 99.21\% & 0.027501 & \textbf{99.26\%} \\
 \hline
1000 & 0.041232 & 99.23\% & 0.030611 & 99.24\%
\end{tabular}
\caption{Result of model selection for mouth and face with the combined parameters for 10, 100, 200, ..., 1000 epochs. Optimal values per part in \textbf{bold}. For reduced dataset.}
\label{table:cv_both_complete_1000_red}
\end{table}

\end{document}